\documentclass[runningheads]{llncs}

\usepackage{eccv}
\usepackage{pdfpages}

\usepackage[table]{xcolor} 
\usepackage{eccvabbrv}

\usepackage{graphicx}
\usepackage{booktabs}

\usepackage[accsupp]{axessibility}  
\usepackage{booktabs}
\usepackage{multirow}
\usepackage{graphicx}
\usepackage{bbm}

\usepackage{hyperref}

\usepackage{orcidlink}
\usepackage{algorithm}
\usepackage{algpseudocode}
\usepackage{adjustbox}
\usepackage{pifont}

\usepackage{amsmath}
\DeclareMathOperator*{\argmax}{arg\,max}

\begin{document}

\title{\textsc{VideoSearch-R1}: Iterative Video Retrieval \\ and Reasoning via Soft Query Refinement}

\titlerunning{VideoSearch-R1}

\author{
Seohyun Lee\inst{1}$^{*}$\orcidlink{0009-0003-4661-3486}
\and
Seoung Choi\inst{1}$^{*}$\orcidlink{0009-0006-4885-4074}
\and
Dohwan Ko\inst{2}$^{*}$\orcidlink{0000-0002-2284-1009}
\and
Jongha Kim\inst{2}\orcidlink{0009-0007-4304-941X}
\and
Hyunwoo J. Kim\inst{1}$^{\dagger}$\orcidlink{0000-0002-2181-9264}
}

\authorrunning{S.~Lee et al.}

\institute{
KAIST, Daejeon, Republic of Korea\\
\email{\{seohyunlee, choisw0823, hyunwoojkim\}@kaist.ac.kr}
\and
Korea University, Seoul, Republic of Korea\\
\email{\{ikodoh, jonghakim\}@korea.ac.kr}
}

\maketitle
{\let\thefootnote\relax\footnotetext{$^{*}$ Equal contribution. \quad $^{\dagger}$ Corresponding author.}}
\begin{abstract}
    As video corpora continue to expand in both scale and task complexity, there is increasing demand for approaches that retrieve relevant videos from large-scale corpora (inter-video reasoning) and subsequently perform fine-grained, query-conditioned tasks (intra-video reasoning) within the retrieved content, such as temporal grounding.
    However, existing approaches typically treat retrieval as a preprocessing step, and consequently, when the initial retrieval fails, there is no mechanism to refine the search, leading to the failure of subsequent fine-grained intra-video reasoning.
    Moreover, while recent agentic frameworks have advanced video understanding, they typically assume that the query-relevant video is already given, focusing exclusively on intra-video reasoning tasks.
    To address these limitations, we propose \textbf{\textsc{VideoSearch-R1}}, an agentic framework for iterative video retrieval and reasoning through multi-turn interaction with a video search engine. 
    Specifically, we introduce \textbf{Soft Query Refinement (SQR)} to refine search query tokens in a continuous latent space rather than rewriting queries in the discrete text space, enabling more efficient and fine-grained adjustments.
    SQR and its reasoning process are trained using Group Relative Policy Optimization (GRPO), guided by task-level reward signals derived from retrieval and downstream tasks. 
    Building upon this, \textsc{VideoSearch-R1} achieves state-of-the-art performance across three datasets on Video Corpus Moment Retrieval (VCMR), iteratively retrieving videos from large-scale corpora, refining search queries, and performing precise query-conditioned temporal grounding within the retrieved content.
    Our analyses show that SQR effectively refines the original query, requiring significantly fewer generated tokens than explicit text-level query refinement. Code and model checkpoints are publicly available at \href{https://mlvlab.github.io/VideoSearch-R1/}{\texttt{mlvlab.github.io/VideoSearch-R1}}.
\end{abstract}

\section{Introduction}
\label{sec:intro}
\begin{figure}[t]
    \centering
    \includegraphics[width=\linewidth]{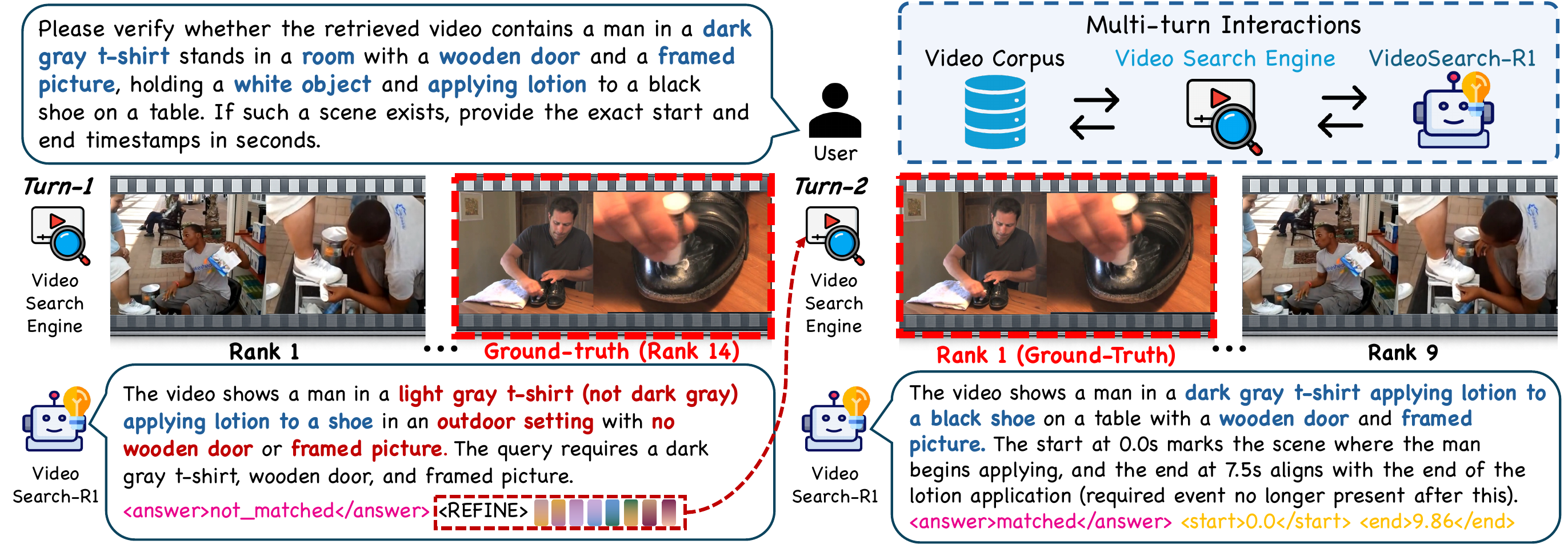}
    \caption{ 
        \textbf{An illustrative example of \textsc{VideoSearch-R1.}}
        As an agentic AI system, \textsc{VideoSearch-R1} enables multi-turn interaction through iterative video retrieval and reasoning, leveraging an external video search engine. 
        This pipeline unifies corpus-level inter-video reasoning (\eg, video retrieval) with intra-video reasoning (\eg, temporal grounding) grounded in the retrieved video.
    }
    \label{fig:teaser}
    \vspace{-8pt}
\end{figure}
With the rapid growth of large-scale video corpora, recent studies have focused on efficiently and accurately retrieving relevant videos given a user query~\cite{luo2021clip4clip,xue2022clip,wu2023cap4video,gorti2022x,liu2022ts2,linmm,liu2025lamra,lee2025captioning,ko2025bidirectional}. 
Although these approaches achieve strong performance on standard video-level retrieval benchmarks through inter-video reasoning, identifying the correct video alone is insufficient for real-world applications.
In practice, users require not only coarse inter-video reasoning but also query-specific intra-video reasoning within the retrieved video.
For example, beyond identifying a relevant video, a system may need to conduct fine-grained reasoning, such as localizing the exact timestamp of a described event, extracting temporally grounded evidence, or performing question answering over the retrieved video.

However, existing pipelines~\cite{hou2021conquer,yoon2022selective,zhang2021video} treat inter-video retrieval as a preprocessing stage prior to intra-video reasoning: inter-video retrieval models~\cite{luo2021clip4clip,xue2022clip,wu2023cap4video,gorti2022x,liu2022ts2,linmm,liu2025lamra,lee2025captioning,ko2025bidirectional} optimize for coarse-grained relevance over a video corpus, whereas intra-video reasoning modules~\cite{yang2023vid2seq,huang2024vtimellm,cheng2024videollama,cao2025flashvtg,yan2025videochat,wang2025adatooler} operate independently within individual videos.
Consequently, such pipelines are limited by their decoupled architecture, where failures in inter-video retrieval propagate to subsequent intra-video reasoning.
These limitations motivate an \emph{iterative video retrieval-and-reasoning} framework within an agentic AI system, where retrieval and reasoning are tightly integrated through an interactive loop. 
Rather than treating inter-video retrieval as a one-shot preprocessing step, the system autonomously retrieves relevant videos, dynamically refines search queries, and performs query-conditioned intra-video reasoning over the retrieved content.
As in Fig.~\ref{fig:teaser}, this framework enables holistic inter-video reasoning across large-scale video corpora as well as fine-grained intra-video reasoning through multi-turn interaction.

Such agentic paradigms have recently shown strong potential in natural language processing through Retrieval-Augmented Generation (RAG), where search engines are treated as external tools~\cite{lewis2020retrieval,asai2023self,jin2025search}.
For example, Search-R1~\cite{jin2025search} introduces a reinforcement learning (RL)-based framework that enables Large Language Models (LLMs) to generate search queries and iteratively refine both queries and reasoning to provide a final answer.
Inspired by these advances, similar efforts have emerged in video understanding, integrating search mechanisms with a core Vision-Language Model (VLM)~\cite{luo2024video,ren2025videorag,wu2023cap4video}.
In particular, video agentic frameworks incorporate external tools, such as object trackers, OCR models, temporal localizers, and video captioners, to facilitate long-form video understanding, where models often struggle to identify salient visual cues among extensive visual tokens.
However, unlike text-based agentic systems that explicitly retrieve external knowledge, most existing video agentic frameworks implicitly assume that the query-relevant video is already known, \ie, bypassing the video retrieval stage.
While effective under this assumption, such designs become suboptimal when users expect the system to dynamically identify and retrieve relevant videos from large-scale corpora prior to intra-video reasoning.

To this end, we propose \textbf{\textsc{VideoSearch-R1}}, an agentic framework that integrates a video search engine to perform iterative video retrieval and reasoning through multi-turn interaction.
The framework iteratively retrieves candidate videos, verifies query-video matching, refines search queries, and performs intra-video reasoning.
For query refinement, instead of explicit text-level query refinement (\ie, hard query refinement), we introduce \textbf{Soft Query Refinement (SQR)}, which generates query representations in the continuous latent space, enabling more efficient and fine-grained refinement.
The soft query is appended to the original query to guide subsequent retrieval and is jointly optimized with the reasoning process via Group Relative Policy Optimization (GRPO)~\cite{shao2024deepseekmath} to maximize task-level rewards.
We train and evaluate \textsc{VideoSearch-R1} on the Video Corpus Moment Retrieval (VCMR)~\cite{chen2024verified} task, which requires the model to first retrieve the relevant video from a corpus given a textual query, and subsequently perform temporal grounding to predict the timestamp within the retrieved video that best corresponds to the query.
\textsc{VideoSearch-R1} achieves state-of-the-art performance across three VCMR benchmarks, ActivityNet-FIG, Charades-FIG, and DiDeMo-FIG, on both inter-video retrieval and intra-video temporal grounding.
Our in-depth analysis shows that the proposed SQR effectively refines the original query to improve retrieval performance while requiring substantially fewer generated tokens than hard query refinement.

\noindent To summarize, our contributions are \textbf{threefold}:
\vspace{-1.0mm}
\begin{itemize}
    \item We propose \textsc{VideoSearch-R1}, an agentic framework for iterative video retrieval and reasoning. 
    It iteratively retrieves candidate videos via a video search engine, verifies query-video matching, refines search queries, and performs intra-video reasoning through multi-turn interaction.
    \item We introduce Soft Query Refinement (SQR), which generates soft query tokens in a continuous latent space for fine-grained refinement, while requiring fewer generated tokens than hard query refinement.
    \item By jointly optimizing inter-video retrieval and intra-video reasoning within \textsc{VideoSearch-R1}, it achieves state-of-the-art performance on three VCMR benchmarks in both video retrieval and temporal grounding.
\end{itemize}
\section{Related Works}

\noindent \textbf{Video retrieval and reranking.}
Video retrieval~\cite{luo2021clip4clip,xue2022clip,wu2023cap4video,gorti2022x,liu2022ts2,linmm,liu2025lamra,lee2025captioning,ko2025bidirectional} is a multi-modal task that retrieves the most relevant video given a text query. 
Early approaches~\cite{luo2021clip4clip,xue2022clip,wu2023cap4video,gorti2022x,liu2022ts2} commonly build upon CLIP~\cite{radford2021learning} by encoding videos and texts with separate encoders and ranking candidates via cosine similarity in a shared embedding space. 
While efficient, this dual-encoder paradigm primarily captures coarse-grained alignment due to the lack of cross-modal interaction.
To address this limitation, recent methods leverage VLMs~\cite{wang2024internvideo2,wang2024qwen2,liu2025lamra,ko2025bidirectional,lee2025captioning} to model fine-grained cross-modal interactions and improve retrieval performance by reranking a fixed top-$K$ set of candidate videos for each query.

\noindent \textbf{Multi-turn reasoning of agentic AI.}
The multi-turn reasoning paradigm in agentic AI has recently attracted attention as an effective strategy for tackling complex analytical problems~\cite{jain2025simpledoc,wang2025vidorag,yao2022react,shen2023hugginggpt}.
In this setting, intermediate actions, such as tool invocation or document retrieval, are dynamically determined to progressively refine the solution. 
This framework has also demonstrated promise in video understanding, where multi-turn reasoning facilitates the interpretation of intricate temporal dependencies and long-range events~\cite{min2024morevqa,wang2025videotree,luo2024video}.
Furthermore, recent works adopt RL methods, including GRPO~\cite{shao2024deepseekmath}, to enable optimization of reasoning trajectories and better align intermediate decisions with downstream objectives~\cite{jin2025search,peiyuan2024agile,wang2025vrag,zhou2025reagent,zhang2025thinking,park2026deepvideo}.
In this work, we introduce \textsc{VideoSearch-R1}, which iteratively retrieves relevant videos, refines the query, verifies query-video alignment, and performs intra-video reasoning through multi-turn interactions.

\noindent \textbf{Soft reasoning in LLMs.}
Recent studies explore soft reasoning to reduce the overhead of explicit text-level chain-of-thought by updating continuous latent states to encode intermediate computations while minimizing long-form text generation~\cite{hao2024training,tan2025think,sunlatent,su2025token,xu2025softcot,li2025latent}. 
For instance, Coconut~\cite{hao2024training} leverages the final hidden state of an LLM as a compact representation of the reasoning state. 
Building upon this line of work, we extend the concept of soft reasoning to query refinement for enhanced video retrieval, and propose soft query refinement (SQR). 
\section{Method}
\label{sec:method}

We propose a video agentic model, \textbf{\textsc{VideoSearch-R1}}, an iterative video retrieval-and-reasoning framework that autonomously retrieves videos, verifies query-video matching, refines search queries, and performs intra-video reasoning. 
We also introduce \textbf{Soft Query Refinement (SQR)}, which produces query representations directly in the continuous latent space, enabling efficient and fine-grained adjustments.
We first provide an overview of \textsc{VideoSearch-R1}, followed by a detailed description of its training and inference procedures.

\subsection{\textsc{VideoSearch-R1} with Soft Query Refinement}
\label{subsec:overall}

\begin{figure*}[t]
    \begin{subfigure}[t]{0.49\linewidth}
        \centering
        \includegraphics[width=\linewidth]{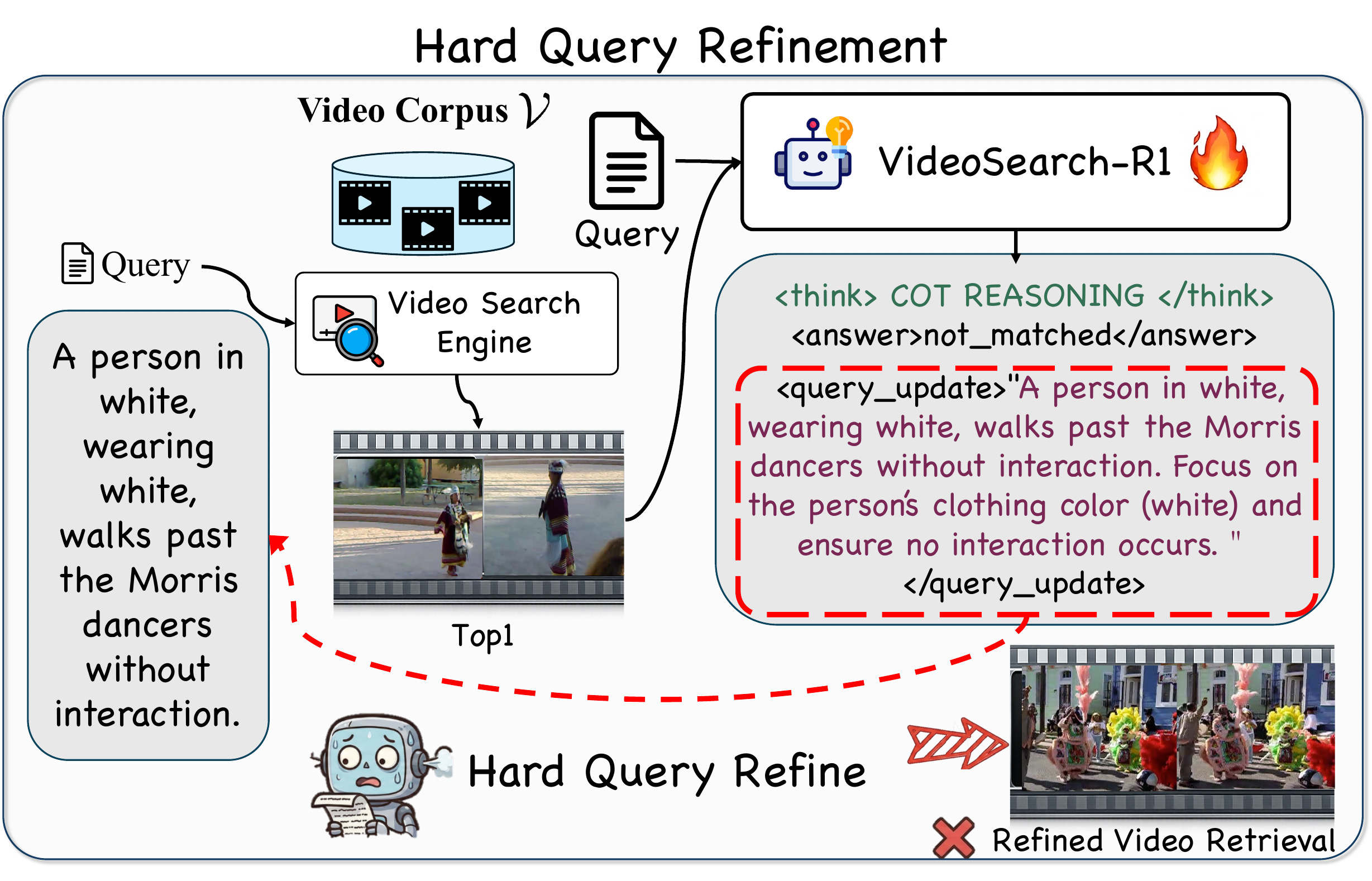}
        \caption{Hard query refinement.}
        \label{fig:sqr1}
    \end{subfigure}
    \begin{subfigure}[t]{0.49\linewidth}
        \centering
        \includegraphics[width=\linewidth]{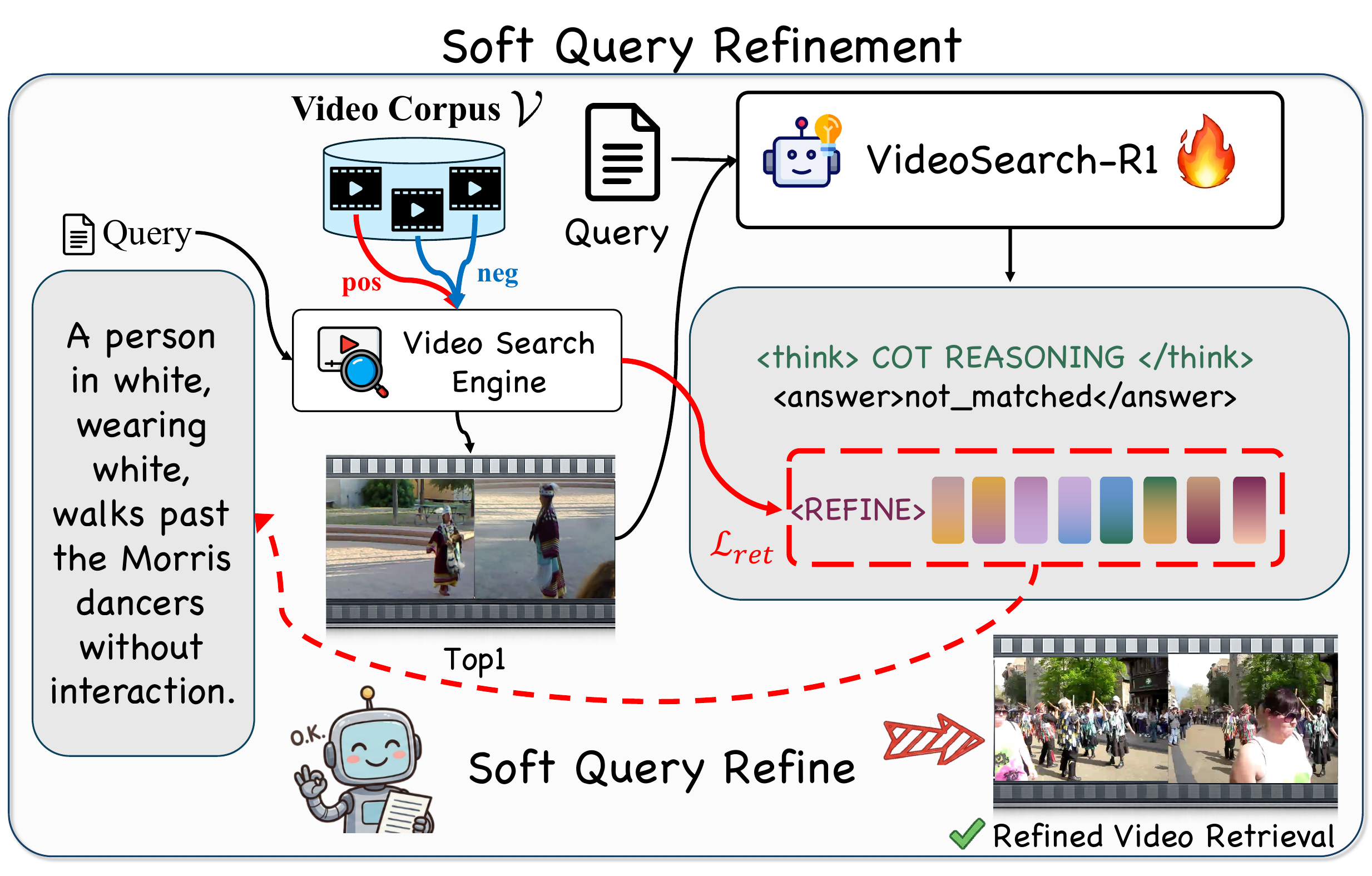}
        \caption{Soft query refinement (\textbf{Ours}).}
        \label{fig:sqr2}
    \end{subfigure}
    \caption{
        \textbf{Comparison between hard query refinement and our Soft Query Refinement (SQR).}
        SQR generates soft query tokens to perform fine-grained adjustments to the original query representation.
        In SQR, the soft query tokens are trained using the InfoNCE objective $\mathcal{L}_\text{ret}$, which provides richer discriminative supervision than the standard next-token prediction used in hard query refinement.
    }
    \label{fig:sqr}
\end{figure*}

Given a user query, \textsc{VideoSearch-R1} iteratively retrieves candidate videos through external search engine calls and verifies their relevance to the query.
Let $t \geq 1$ denote the current turn of the iterative video retrieval-and-reasoning loop, initialized with the original user query $q_1$.
At turn $t$, \textsc{VideoSearch-R1} invokes a video search engine $\mathcal{R}$, a cross-modal dense embedding retriever (Qwen3-VL-Embedding-2B~\cite{li2026qwen3}), using the query $q_t$.
The search engine returns the top-1 candidate video $v_t$ from the video corpus $\mathcal{V}$ based on global similarity as:
\begin{equation}
    v_t = \mathcal{R}(q_t) = \argmax_{v \in \mathcal{V}} f(q_t)^\top f(v),
\end{equation}
where $f$ represents the encoder of the video search engine $\mathcal{R}$.
However, video-level retrieval does not guarantee fine-grained semantic alignment with the query.
To refine retrieval at turn $t$ through iterative video retrieval and reasoning, \textsc{VideoSearch-R1} performs a verification step with the reasoning process.
Given the current query $q_t$ and the retrieved video $v_t$, the model evaluates whether the video content satisfies the intended temporal semantics.
It then generates a matching indicator $y_t^\text{ret} \in \{\texttt{`match'}, \texttt{`not match'}\}$, representing whether $v_t$ aligns with $q_t$, along with the corresponding intermediate reasoning trace $r_t$.

If the retrieved result is deemed a mismatch (\ie, $y_t^\text{ret}$ = \texttt{`not match'}), the model performs SQR, which autoregressively generates a fixed number of soft query tokens $q_t^\text{soft} \in \mathbb{R}^{N \times D}$ as continuous latent embeddings, where $N$ is the number of soft query tokens and $D$ is the hidden dimension.
Specifically, during autoregressive decoding, the hidden state corresponding to the previously generated token is projected through a linear layer and used directly as the input embedding for the next token.
After generating $N$ soft query tokens $q_t^\text{soft}$, we append them to the original query $q_1$ to form the refined query for the next turn, defined as $q_{t+1} = [q_1 \| q_t^\text{soft}]$.
This refined query is then used to re-invoke the search engine.
As in Fig.~\ref{fig:sqr}, unlike hard query refinement based on explicit text-level rewriting (Fig.~\ref{fig:sqr1}), SQR enables more fine-grained adjustments to the query representation while requiring fewer generated tokens (Fig.~\ref{fig:sqr2}).


\begin{figure*}[t]
    \centering
    \includegraphics[width=\linewidth]{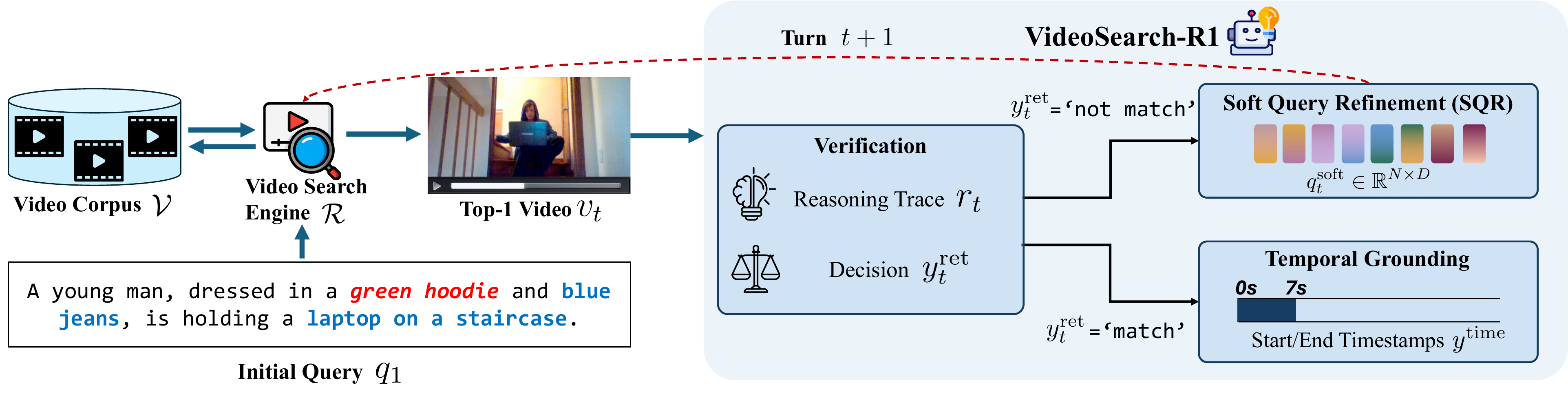}
    \caption{
        \textbf{Iterative video retrieval and reasoning of \textsc{VideoSearch-R1}.}
        Given an initial query $q_1$, \textsc{VideoSearch-R1} retrieves the top-1 video from a corpus via a video search engine and performs verification, producing a reasoning trace $r_t$ and a matching decision $y_t^\text{ret}$. 
        If $y_t^\text{ret} =$ \texttt{`not match'}, the model performs SQR by generating soft query tokens $q_t^\text{soft} \in \mathbb{R}^{N \times D}$ to construct a refined query $q_{t+1} = [q_1 \| q_t^\text{soft}]$. 
        If matched, the model conducts temporal grounding to predict the start and end timestamps $y^\text{time}$.
    }
    \label{fig:main}
\end{figure*}
This iterative retrieval and reasoning continues until the $k$-th turn, where either a valid match is identified or a predefined maximum number of turns $T$ is reached.
When a valid match is identified (\ie, $y_t^\text{ret} = \texttt{`match'}$), the model proceeds to intra-video reasoning to predict the precise temporal boundaries $y^\text{time}$ (\ie, start and end timestamps) of the query-relevant segment within the retrieved video.
If no valid match is found within the allowed iterations, the episode is treated as a failure case.
The overall framework is illustrated in Fig.~\ref{fig:main}, and the template for each interaction turn is presented in Tab.~\ref{tab:template}.

\subsection{Training Procedure}
\label{subsec:training}

\begin{table}[t]
    \caption{
        \textbf{Template of a single turn within the multi-turn interaction of \textsc{VideoSearch-R1}.}
        This process iterates until either the maximum number of turns is reached or the model verifies that the retrieved video matches the query.
    }
    \label{tab:template}
    \centering
    \begin{tabular}{p{0.98\linewidth}}
        \toprule
        {\raggedright
        \textcolor{blue}{\textbf{System prompt:}} You are a video retrieval assistant. Your task is to analyze a retrieved video against the user query. Inside \textcolor{teal}{\texttt{<think>...</think>}}, perform a step by step comparison between the query requirements and the visible evidence in the video. Identify whether a scene corresponding to the query appears in the video and determine the exact time span where it occurs. If a scene corresponding to the query appears in the video, output strictly in the following format: \textcolor{teal}{\texttt{<think>...</think>}} \textcolor{magenta}{\texttt{<answer>matched</answer>}} \textcolor{orange}{\texttt{<start>...</start> <end>...</end>}}\textcolor{violet}{\texttt{<REFINE>}}. Even if matched, you must still append the special token(s). \textcolor{violet}{\texttt{<REFINE>}} at the very end to allow further latent refinement. If no scene corresponding to the query appears in the video, output strictly: \textcolor{red}{\texttt{<answer>not matched</answer>}}\textcolor{violet}{\texttt{<REFINE>}}. In this case, the special token(s) are required to initiate a latent query update. You must always append the special token(s) \textcolor{violet}{\texttt{<REFINE>}} at the very end of the output. Do not invent details beyond what is visible. Be concise inside \textcolor{teal}{\texttt{<think>...</think>}}. Do not output anything outside the specified tags.  \par} \\
        \midrule
        \midrule
        \textcolor{blue}{\textbf{User:}} \texttt{[user query]}, \texttt{[retrieved video]} \\
        \midrule
        \midrule
        {\raggedright
        \textcolor{blue}{\textbf{Assistant (match):}} \textcolor{teal}{\texttt{<think>...</think>}} \textcolor{magenta}{\texttt{<answer>matched</answer>}} \textcolor{orange}{\texttt{<start>...</start> <end>...</end>}}\textcolor{violet}{\texttt{<REFINE>}} \par} \\
        \midrule
        {\raggedright
        \textcolor{blue}{\textbf{Assistant (mismatch):}} \textcolor{teal}{\texttt{<think>...</think>}} \textcolor{red}{\texttt{<answer>not matched</answer>}} \textcolor{violet}{\texttt{<REFINE>}} \par} \\
        \bottomrule
    \end{tabular}
\end{table}

To optimize the iterative video retrieval-and-reasoning framework, we adopt a standard two-stage training pipeline widely used in video reasoning models~\cite{feng2025video,feng2025onethinker,zhang2025thinking}. 
In the first stage, we perform Supervised Fine-Tuning (SFT) to initialize \textsc{VideoSearch-R1} with a structured reasoning template and encourage the generation of meaningful soft query tokens to improve video retrieval.
The second stage applies RL-based policy optimization via GRPO~\cite{shao2024deepseekmath}, enabling the model to explore diverse reasoning trajectories and reinforce high-reward behaviors, thereby enhancing both inter-video retrieval accuracy and intra-video reasoning quality.
We instantiate this training paradigm on the Video Corpus Moment Retrieval (VCMR) task, which naturally aligns with our unified objective: given a textual query, the model must first retrieve the relevant video from a large-scale corpus and subsequently predict the precise temporal boundaries within the retrieved video that best correspond to the query.

\noindent \textbf{Stage 1: SFT cold start.}
In the SFT stage, we initialize \textsc{VideoSearch-R1} with Qwen3-VL-2B-Instruct~\cite{bai2025qwen3} and train it to follow a structured reasoning pattern while developing soft query generation capabilities within the iterative video retrieval-and-reasoning pipeline.
Specifically, we supervise a reasoning trace $r$ and two output variables, $y^\text{ret}$ and $y^\text{time}$, corresponding to query-video matching verification and precise timestamp prediction, respectively.
The reasoning trace and output variables are optimized using the following objectives:
\begin{equation}
    \mathcal{L}_\text{verif} = -\log P(r, y^\text{ret}|q,v), \:\: 
    \mathcal{L}_\text{time} = -\log P(y^\text{time}|q,v,r,y^\text{ret}).
\end{equation}

To obtain high-quality Chain-of-Thoughts (CoT) annotations of the reasoning trace $r$, we leverage a powerful VLM, Qwen3-VL-30B-A3B-Thinking~\cite{bai2025qwen3}. 
We sample 2K query-video pairs from the VCMR dataset, consisting of 1K matching cases (where the search engine retrieves the correct video) and 1K negative cases (where retrieval fails). 
For positive (matching) query-video pairs, the model is trained to generate both an explanation of the semantic alignment between the query and the retrieved video, $\mathcal{L}_\text{verif}$, and a justification for the ground-truth temporal boundaries, $\mathcal{L}_\text{time}$.
For negative pairs, supervision is applied only to $\mathcal{L}_\text{verif}$, encouraging the model to explain the semantic mismatch between the query and the video.
Subsequently, the model is encouraged to refine the query via soft query generation and re-invoke the search engine in the next turn.

While CoT traces can be directly annotated from ground-truth query-video pairs, soft query tokens lack explicit supervision.
To address this, we optimize soft queries using a contrastive objective based on InfoNCE~\cite{oord2018representation} as in Fig.~\ref{fig:sqr2}.
Concretely, after the model autoregressively generates $N$ soft query tokens $q^\text{soft}$, these tokens are appended to the original query and optimized to maximize similarity with the ground-truth video $v$ while minimizing similarity with negative videos $\mathcal{V}^\text{neg}$ in the search engine's embedding space, formulated as:
\begin{equation}
\resizebox{\linewidth}{!}{%
    $\mathcal{L}_\text{ret} = -\log \left( \frac{\exp\left( f\left(\left[q_1 \| q^\text{soft}\right]\right)^\top f\left(v\right) \right)}{\exp\left( f\left(\left[q_1 \| q^\text{soft}\right]\right)^\top f\left(v\right) \right) + \sum_{v^- \in \mathcal{V}^\text{neg}} \exp\left(f\left(\left[q_1 \| q^\text{soft}\right]\right)^\top f\left(v^-\right) \right)} \right).$
    \label{eq:infonce}
}
\end{equation}
By explicitly incorporating negative video information through Eq.~\eqref{eq:infonce}, this contrastive objective provides richer and more discriminative supervision for SQR than conventional next-token prediction used to train hard query refinement.

As a result, the overall SFT objective is defined as:
\begin{equation}
    \mathcal{L}_\text{SFT} = \mathcal{L}_\text{verif} + \mathcal{L}_\text{ret} + \mathbbm{1}_{y^\text{ret} = \texttt{`match'}}(\mathcal{L}_\text{time}).
\end{equation}

\noindent \textbf{Stage 2: Training via GRPO.}
Once the model acquires structured reasoning patterns and soft query generation capabilities, we proceed to the second stage to optimize the model using GRPO. 
While SFT provides a stable initialization, it does not explicitly optimize the interaction between retrieval and reasoning. 
We therefore employ RL to explore improved reasoning trajectories and more effective query refinement strategies.
To align policy optimization with the objectives of the interleaved framework, we design four complementary reward signals: \emph{format}, \emph{verification}, \emph{retrieval}, and \emph{temporal grounding}. 

The format reward $R^\text{format}$ encourages the model to follow predefined reasoning and soft query structures, enforcing compliance with the template in Tab.~\ref{tab:template}, which reflects the iterative video retrieval and reasoning process.
A reward of 1 is assigned if the model output strictly follows the required format, \eg, the reasoning process enclosed in \texttt{<think>...</think>}, the matching result in \texttt{<answer>...</answer>}, and the predicted timestamps in \texttt{<start>...</start>} and \texttt{<end>...</end>}, and 0 otherwise.
The verification reward $R^\text{verif}$ supervises the query-video matching verification. 
A reward of 1 is assigned if the model correctly determines whether the video matches the query, and a reward of 0 is assigned otherwise.
To supervise the soft query tokens during RL, we reuse the retrieval objective $\mathcal{L}_\text{ret}$ from the SFT stage and define the corresponding reward as $R^\text{ret} = \exp(-\mathcal{L}_\text{ret})$, encouraging improved retrieval performance.
Finally, the temporal grounding reward $R^\text{time}$ promotes accurate moment localization within the retrieved content by directly calculating the IoU between the predicted and ground-truth timestamps, reinforcing intra-video reasoning quality.

Overall, the final reward for the $i$-th sample is defined as:
\begin{equation}
    R_i = R_i^\text{format} + R_i^\text{verif} + R_i^\text{ret} + \mathbbm{1}_{y^\text{ret} = \texttt{`match'}}(R_i^\text{time}).
\end{equation}
The advantage $A_i$ is then computed by normalizing rewards within each group:
\begin{equation}
    A_i = \frac{R_i - \text{mean}(\{R_j\})}{\text{std}(\{R_j\})}.
\end{equation}
We adopt GRPO as the RL algorithm to train \textsc{VideoSearch-R1}, propagating reward signals across both inter-video retrieval and intra-video reasoning, resulting in a holistic optimization of iterative video retrieval and reasoning.


\subsection{Inference via Multi-Turn Interaction}
\label{subsec:inference}

\textsc{VideoSearch-R1} performs multi-turn interaction through iterative video retrieval and reasoning, consisting of external video search, retrieved video verification, soft query refinement, and temporal grounding within the selected video.
At each turn, the model first assesses the semantic alignment between the current query and the retrieved video, given the original query. 
If a mismatch is identified, it generates $N$ soft query tokens, and these tokens are appended to the original query embedding sequence and fed into the video search engine $\mathcal{R}$. 
The search engine then re-ranks candidate videos conditioned on the refined query and returns the top-1 video. 
The newly retrieved video is incorporated into the context for the subsequent verification step.
This multi-turn process continues until either the model verifies a correct match and produces a final temporal grounding prediction or the maximum number of retrieval turns $T$ is reached, in which case the episode is considered a failure.
\section{Experiments}
\label{sec:exp}

\noindent \textbf{Benchmarks and datasets.}
To evaluate the joint inter-video retrieval and intra-video reasoning capabilities of \textsc{VideoSearch-R1}, we adopt the recently introduced Video Corpus Moment Retrieval (VCMR) task~\cite{chen2024verified}, a challenging benchmark for corpus-level temporal grounding.
VCMR requires the model to first retrieve the video relevant to a given textual query from the corpus (video retrieval) and then localize the precise start and end timestamps of the query within the retrieved video (temporal grounding).
We conduct experiments on three VCMR benchmarks: ActivityNet-FIG, DiDeMo-FIG, and Charades-FIG.

\noindent \textbf{Evaluation metrics.}
We report performance on VCMR and its subtask, video retrieval (VR). 
For VR, we adopt Recall@$K$ (R@$K$) with $K \in \{1, 5, 10, 100\}$. 
For VCMR, we evaluate end-to-end performance using R@$K$ under different temporal overlap thresholds, specifically IoU $\in \{0.3, 0.5, 0.7\}$ (denoted as IoU/R@1). 
A prediction is considered correct if the model identifies the ground-truth video and the IoU between the predicted and ground-truth temporal spans exceeds the specified threshold.
Failure to retrieve the correct video yields a zero score for the VCMR metric. 
We additionally evaluate verification (VER) performance by measuring the accuracy of predicting \texttt{`match'} or \texttt{`not match'}.

\noindent \textbf{Implementation details.}
We employ Qwen3-VL-Embedding-2B~\cite{li2026qwen3} as the video search engine, which retrieves the top-1 video given a textual query from a large-scale video corpus. 
\textsc{VideoSearch-R1} is fine-tuned based on Qwen3-VL-2B-Instruct~\cite{bai2025qwen3}.
During training, the total number of visual tokens is set to 4,096 by sampling videos at 1 FPS with up to 64 frames. 
We use $N=8$ soft query tokens.
In Stage 1 (SFT), we fine-tune the model for 2K steps over 2K samples per dataset using AdamW with a learning rate of 2e-5. 
In Stage 2 (RL), we initialize from the SFT checkpoint and optimize with AdamW using a learning rate of 5e-7, weight decay of 0.01, and a maximum gradient norm of 1.0. 
We set the KL coefficient to $\beta=0.01$, the rollout size to $G=8$, and the decoding temperature to 1.0.
We set the maximum number of inference turns to $T=2$.


\noindent \textbf{Baselines.}
To ensure a fair comparison, we introduce two baselines built upon the same backbone VLM, Qwen3-VL-2B-Instruct~\cite{bai2025qwen3}, with an identical number of parameters. 
The first baseline uses the zero-shot (ZS) model, while the second is fine-tuned (FT) on the same training dataset as \textsc{VideoSearch-R1}.
The fine-tuned baseline is trained to directly predict whether a retrieved video matches the query and to estimate its temporal boundaries, but it does not generate soft query tokens for refinement.
We also apply the same multi-turn inference procedure as in \textsc{VideoSearch-R1}. 
Specifically, if the model determines that the top-1 retrieved video does not match the query, it sequentially evaluates the top-2, top-3, and subsequent candidates returned by the search engine. 
\vspace{-3.5mm}

\subsection{Main Results}

\begin{table*}[t]
    \caption{
        \textbf{Results on VCMR, VER, and VR.}
    }
    \label{tab:main}

    \centering
    \setlength{\tabcolsep}{3pt}
    \begin{adjustbox}{width=\textwidth}
    \begin{tabular}{ll|ccc|c|cccc}
        \toprule
        \multirow{2}{*}{\textbf{Dataset}} & \multirow{2}{*}{\textbf{Method}} & \multicolumn{3}{c|}{\textbf{VCMR}} & \textbf{VER} & \multicolumn{4}{c}{\textbf{VR}} \\
        \cmidrule(lr){3-5}\cmidrule{6-6}\cmidrule(lr){7-10}
        & & 0.3/R@1 & 0.5/R@1 & 0.7/R@1 & Acc & R@1 & R@5 & R@10 & R@100 \\        
        \midrule
        
        \multirow{5}{*}{Charades-FIG}
        & CONQUER~\cite{hou2021conquer} & -- & 1.2 & 0.7 & -- & 2.8 & 9.0 & 14.1 & 51.7 \\
        & SQuiDNet~\cite{yoon2022selective} & -- & 2.6 & 0.9 & -- & 11.7 & 33.9 & 44.0 & 51.7 \\
        \cmidrule{2-10}
        & Qwen3-VL-2B (ZS) & 12.2 & 7.2 & 2.9 & 30.0 & \multirow{2}{*}{21.6} & \multirow{2}{*}{41.8} & \multirow{2}{*}{51.5}& \multirow{2}{*}{84.2} \\
        & Qwen3-VL-2B (FT) & 12.9 & 10.4 & 7.2 & 74.7 &  &  &  &  \\
        \cmidrule{2-10}

        \rowcolor[HTML]{F0F8FF}
        & \textbf{\textsc{VideoSearch-R1}} & \textbf{16.5} & \textbf{13.4} & \textbf{8.2} & \textbf{75.7} & \textbf{24.6} & \textbf{42.9} & \textbf{52.2} & \textbf{84.7} \\
        \midrule
 
        \multirow{5}{*}{DiDeMo-FIG}
        & CONQUER~\cite{hou2021conquer} & -- & 5.5 & 3.7 & -- & 14.8 & 40.4 & 54.0 & 91.5 \\
        & SQuiDNet~\cite{yoon2022selective} & -- & 2.9 & 0.5 & -- & 16.9 & 44.6 & 59.3 & 91.5 \\
        \cmidrule{2-10}
        & Qwen3-VL-2B (ZS)
        & 22.0 & 10.6 & 4.0 & 62.8 
        & \multirow{2}{*}{54.8} 
        & \multirow{2}{*}{79.3} 
        & \multirow{2}{*}{85.6} 
        & \multirow{2}{*}{97.0} \\
        & Qwen3-VL-2B (FT) 
        & 23.6 & 22.1 & 16.7 & 73.1 
        &  &  &  &  \\
        \cmidrule{2-10}

        \rowcolor[HTML]{F0F8FF} & \textbf{\textsc{VideoSearch-R1}} & \textbf{33.3} & \textbf{30.2} & \textbf{19.7} & \textbf{74.6} & \textbf{59.0} & \textbf{82.0} & \textbf{87.8} & \textbf{97.5} \\
        \midrule
 
        \multirow{5}{*}{ActivityNet-FIG}
        & CONQUER~\cite{hou2021conquer} & -- & 3.0 & 1.6 & -- & 13.5 & 36.4 & 50.0 & 89.3 \\
        & SQuiDNet~\cite{yoon2022selective} & -- & 4.7 & 2.1 & -- & 32.6 & 79.9 & 87.9 & 89.3 \\
        \cmidrule{2-10}
        & Qwen3-VL-2B (ZS)
        & 17.2 & 10.1 & 5.8 & 63.0 
        & \multirow{2}{*}{55.1} 
        & \multirow{2}{*}{78.8} 
        & \multirow{2}{*}{86.7} 
        & \multirow{2}{*}{98.1} \\
        
        & Qwen3-VL-2B (FT) 
        & 29.1 & 19.2 & 11.4 & 83.1 
        &  &  &  &  \\
        \cmidrule{2-10}

        \rowcolor[HTML]{F0F8FF}
        & \textbf{\textsc{VideoSearch-R1}} & \textbf{33.8} & \textbf{22.3} & \textbf{12.3} & \textbf{83.3} & \textbf{61.1} & \textbf{81.7} & \textbf{88.5} & \textbf{98.4} \\
        
        \bottomrule
    \end{tabular}
    \end{adjustbox}
\end{table*}
In Tab.~\ref{tab:main}, we evaluate \textsc{VideoSearch-R1} on Charades-FIG, DiDeMo-FIG, and ActivityNet-FIG. 
Qwen3-VL-2B (ZS) and Qwen3-VL-2B (FT) achieve identical video retrieval (VR) results, as neither baseline updates the query representation during multi-turn inference. 
In contrast, \textsc{VideoSearch-R1} iteratively refines the query representation via SQR, resulting in substantial improvements in VR despite using the same search engine. 
For example, on ActivityNet-FIG, R@1 improves by 6.0, underscoring the importance of iterative query refinement for accurate retrieval in multi-turn interaction.
Furthermore, \textsc{VideoSearch-R1} substantially outperforms both zero-shot and fine-tuned baselines in verification accuracy (VER) and temporal grounding for VCMR.
Notably, \textsc{VideoSearch-R1} improves 0.3/R@1 by 9.7 on DiDeMo-FIG compared to Qwen3-VL-2B (FT). 
These results demonstrate the effectiveness of jointly optimizing retrieval and reasoning within an agentic framework through RL-based policy learning.


\subsection{Ablation Studies}

\begin{table*}[t]
    \caption{
        \textbf{Ablation studies of training stages on DiDeMo-FIG.}
    }
    \label{tab:stage}

    \centering
    \setlength{\tabcolsep}{3pt}
    \begin{adjustbox}{width=\textwidth}
    \begin{tabular}{l|c c c|c|c c c c}
        \toprule
        \multirow{2}{*}{\textbf{Method}} & \multicolumn{3}{c|}{\textbf{VCMR}} & \textbf{VER} & \multicolumn{4}{c}{\textbf{VR}} \\
        \cmidrule(lr){2-4}\cmidrule{5-5}\cmidrule(lr){6-9}
        & 0.3/R@1 & 0.5/R@1 & 0.7/R@1 & Acc & R@1 & R@5 & R@10 & R@100 \\        
        \midrule
        Qwen3-VL-2B (ZS) & 22.0 & 10.6 & 4.0 & 62.8 & 54.8 & 79.3 & 85.6 & 97.0 \\
        \cmidrule{1-9}
        \textbf{\textsc{VideoSearch-R1}} (Stage1) & 20.4 & 18.7 & 14.0 & 66.0 & 57.4 & 80.6 & 86.8 & 97.3 \\
        \rowcolor[HTML]{F0F8FF}
        \textbf{\textsc{VideoSearch-R1}} (Stage1 + Stage2) & \textbf{33.3} & \textbf{30.2} & \textbf{19.7} & \textbf{74.6} & \textbf{59.0} & \textbf{82.0} & \textbf{87.8} & \textbf{97.5} \\  
        \bottomrule
    \end{tabular}
    \end{adjustbox}
\end{table*}
\noindent \textbf{Effect of training stages.}
In Tab.~\ref{tab:stage}, compared to the zero-shot model, the SFT cold start (Stage 1) establishes a strong foundation by enforcing the prescribed reasoning template and equipping the model with basic SQR capabilities, resulting in substantial improvements in R@1 for VR. 
However, the gains in temporal grounding on VCMR remain marginal.
In contrast, Stage 2 (SFT + RL) yields pronounced improvements after applying GRPO.
This discrepancy suggests that while SFT primarily enhances structural reasoning patterns, it is less effective at improving genuine temporal reasoning, which is further strengthened through subsequent RL training.

\begin{table*}[t]
    \caption{
        \textbf{Ablation studies of reward design on DiDeMo-FIG.}
    }
    \label{tab:reward}

    \centering
    \setlength{\tabcolsep}{5pt}
    \begin{adjustbox}{width=\textwidth}
    \begin{tabular}{l|c c c|c c c|c|c c c c}
        \toprule
        \multirow{2}{*}{\textbf{Stage}} & \multicolumn{3}{c|}{\textbf{Rewards}} & \multicolumn{3}{c|}{\textbf{VCMR}} & \textbf{VER} & \multicolumn{4}{c}{\textbf{VR}} \\
        \cmidrule(lr){2-4}\cmidrule(lr){5-7}\cmidrule{8-8}\cmidrule(lr){9-12}
        & $R^\text{ret}$ & $R^\text{verif}$ & $R^\text{time}$ & 0.3/R@1 & 0.5/R@1 & 0.7/R@1 & Acc & R@1 & R@5 & R@10 & R@100 \\        
        \midrule
        Stage 1 & & & & 20.4 & 18.7 & 14.0 & 66.0 & 57.4 & 80.6 & 86.8 & 97.3 \\
        \midrule
        Stage 1 + Stage 2 & \ding{52} & & & 18.6 & 17.3 & 13.0 & 65.0 & 58.0 & 81.4 & 87.4 & 97.5 \\
        Stage 1 + Stage 2 & \ding{52} & \ding{52} & & 19.3 & 17.3 & 13.3 & \textbf{75.0} & \textbf{59.7} & \textbf{82.6} & \textbf{88.3} & \textbf{97.7} \\
        \rowcolor[HTML]{F0F8FF}
        Stage 1 + Stage 2 & \ding{52} & \ding{52} & \ding{52} & \textbf{33.3} & \textbf{30.2} & \textbf{19.7} & 74.6 & 59.0 & 82.0 & 87.8 & 97.5 \\
        \bottomrule
    \end{tabular}
    \end{adjustbox}
\end{table*}
\noindent \textbf{Ablation studies on reward design.}
To examine the individual contributions of each reward signal during the RL stage, we conduct an ablation study on $R^\text{ret}$, $R^\text{verif}$, and $R^\text{time}$, in Tab.~\ref{tab:reward}.
First, introducing the retrieval reward $R^\text{ret}$ improves VR performance, suggesting that this signal encourages SQR to produce query representations that are better aligned with the corresponding video embeddings.
Additionally, incorporating the verification reward $R^\text{verif}$ improves the model's ability to assess semantic consistency between the retrieved video and the query, resulting in substantial improvements in verification accuracy and more reliable retrieval decisions.
Finally, adding the temporal grounding reward $R^\text{time}$ significantly improves temporal localization by promoting more precise reasoning and boundary prediction, with a slight trade-off in VER and VR.
Overall, these results demonstrate that each reward component targets a distinct stage of the iterative retrieval-and-reasoning pipeline, and that their combination enables holistic optimization across structural formatting, retrieval alignment, verification reliability, and fine-grained temporal grounding.

\subsection{Analysis}

\begin{figure*}[t]
    \begin{minipage}[t]{0.48\textwidth}
        \centering
        \includegraphics[width=0.9\linewidth]{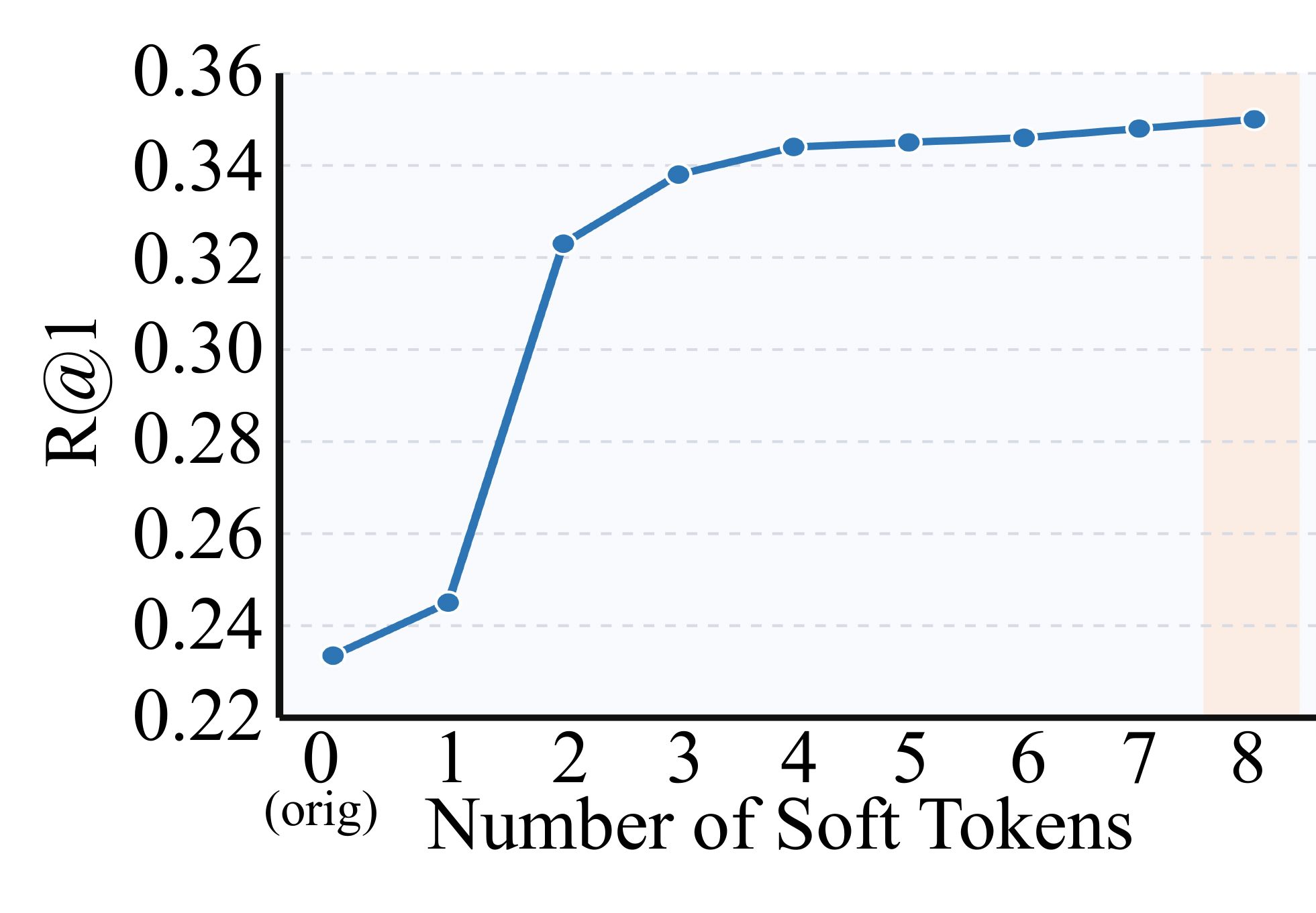}
        \caption{
            \textbf{Effect of the number of soft tokens.}
            R@1 is computed over samples with refined queries.
        }
        \label{fig:num_soft}
    \end{minipage}
    \hfill 
    \begin{minipage}[t]{0.48\textwidth}
        \centering
        \includegraphics[width=0.9\linewidth]{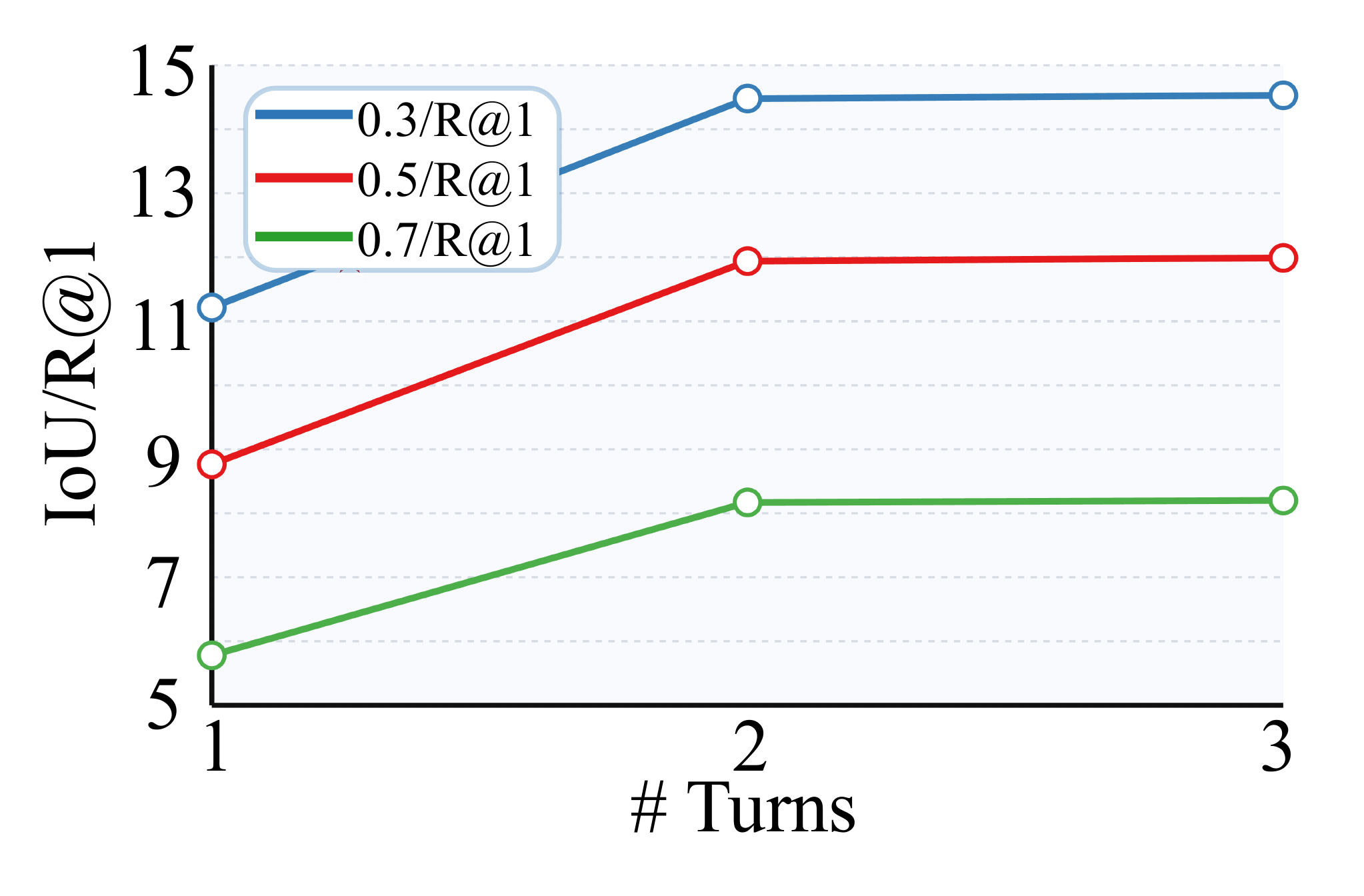}
        \vspace{0.1mm}
        \caption{
            \textbf{Effect of multi-turn inference.}
            The performance on VCMR saturates at $T = 3$.
        }
        \label{fig:turn}
    \end{minipage}
\end{figure*}

\begin{figure}[t]
    \centering
    \includegraphics[width=\linewidth]{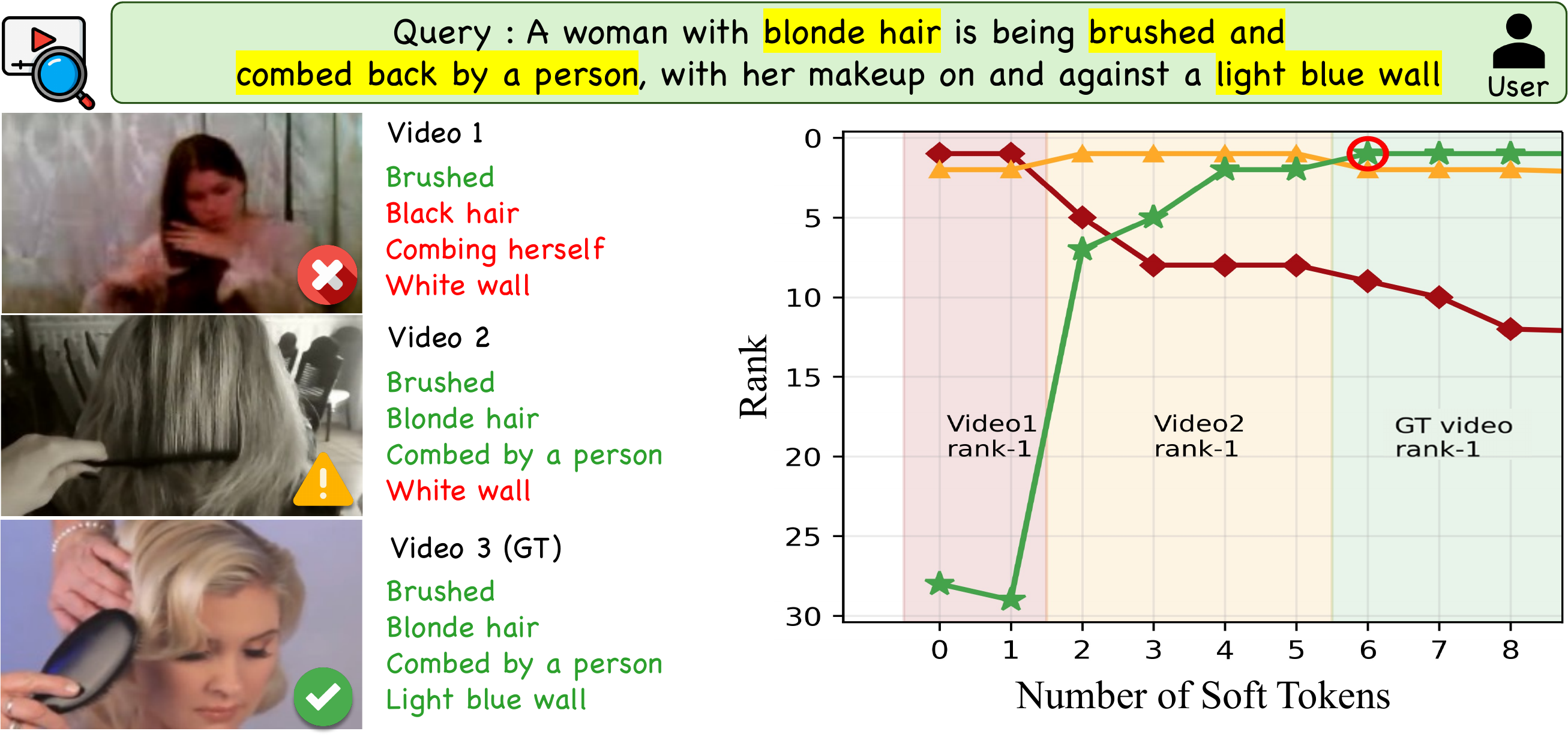}
    \caption{
        \textbf{Changes in the retrieved video as the number of soft tokens increases.}
        The rank of the ground-truth video gradually improves as soft tokens are appended, capturing increasingly fine-grained semantics.
    }
    \label{fig:qual_rank}
    \vspace{-5.5mm}
\end{figure}

\noindent \textbf{Analysis of SQR.}
We first present an in-depth analysis of SQR by examining how retrieval performance evolves as the number of soft query tokens increases. 
In Fig.~\ref{fig:num_soft}, as additional soft query tokens are appended, the average R@1 consistently improves, underscoring that the soft query tokens incrementally refine the original query.
Fig.~\ref{fig:qual_rank} illustrates a qualitative example where the rank of the ground-truth video gradually improves as soft tokens are appended to the original query. 
Without soft query tokens, the video search engine retrieves an incorrect video (video 1), capturing only the coarse concept `a woman brushing'. 
As additional soft tokens are introduced, the retrieval results begin to capture more specific attributes, such as `blonde hair' and `combed by a person', in the video 2. 
Finally, with eight soft tokens, the search engine successfully retrieves the ground-truth video (video 3) at rank 1 by further capturing the context `light blue wall'.
These findings indicate that SQR effectively steers the continuous query representation toward the target video embedding, thereby improving retrieval accuracy and establishing a stronger foundation for intra-video temporal grounding.

\begin{table*}[t]
    \caption{
        \textbf{Quantitative results of hard query refinement (HQR) and our soft query refinement (SQR) on ActivityNet-FIG}
    }
    \label{tab:sqr}

    \centering
    \setlength{\tabcolsep}{3pt}
    \begin{adjustbox}{width=\textwidth}
    \begin{tabular}{l|c c c|c|c c c c|c}
        \toprule
        \multirow{2}{*}{\textbf{Method}} & \multicolumn{3}{c|}{\textbf{VCMR}} & \textbf{VER} & \multicolumn{4}{c|}{\textbf{VR}} & \multirow{2}{*}{\textbf{$\#$ tokens}} \\
        \cmidrule(lr){2-4}\cmidrule{5-5}\cmidrule(lr){6-9}
        & 0.3/R@1 & 0.5/R@1 & 0.7/R@1 & Acc & R@1 & R@5 & R@10 & R@100 \\        
        \midrule

        Qwen3-VL-2B (ZS) & 17.2 & 10.1 & 5.8 & 63.0 & 55.1 & 78.8 & 86.7 & 98.1 & - \\
        \midrule

        \textbf{\textsc{VideoSearch-R1}} + HQR & 33.2 & \textbf{22.3} & 11.9 & 82.2 & 57.6 & 77.2 & 85.4 & 97.8 & 26.8 \\
        \rowcolor[HTML]{F0F8FF}
        \textbf{\textsc{VideoSearch-R1}} + SQR & \textbf{33.8} & \textbf{22.3} & \textbf{12.3} & \textbf{83.3} & \textbf{61.1} & \textbf{81.7} & \textbf{88.5} & \textbf{98.4} & \textbf{8.0} \\        
        \bottomrule
    \end{tabular}
    \end{adjustbox}
\end{table*}
\begin{figure}[t]
    \centering
    \includegraphics[width=\linewidth]{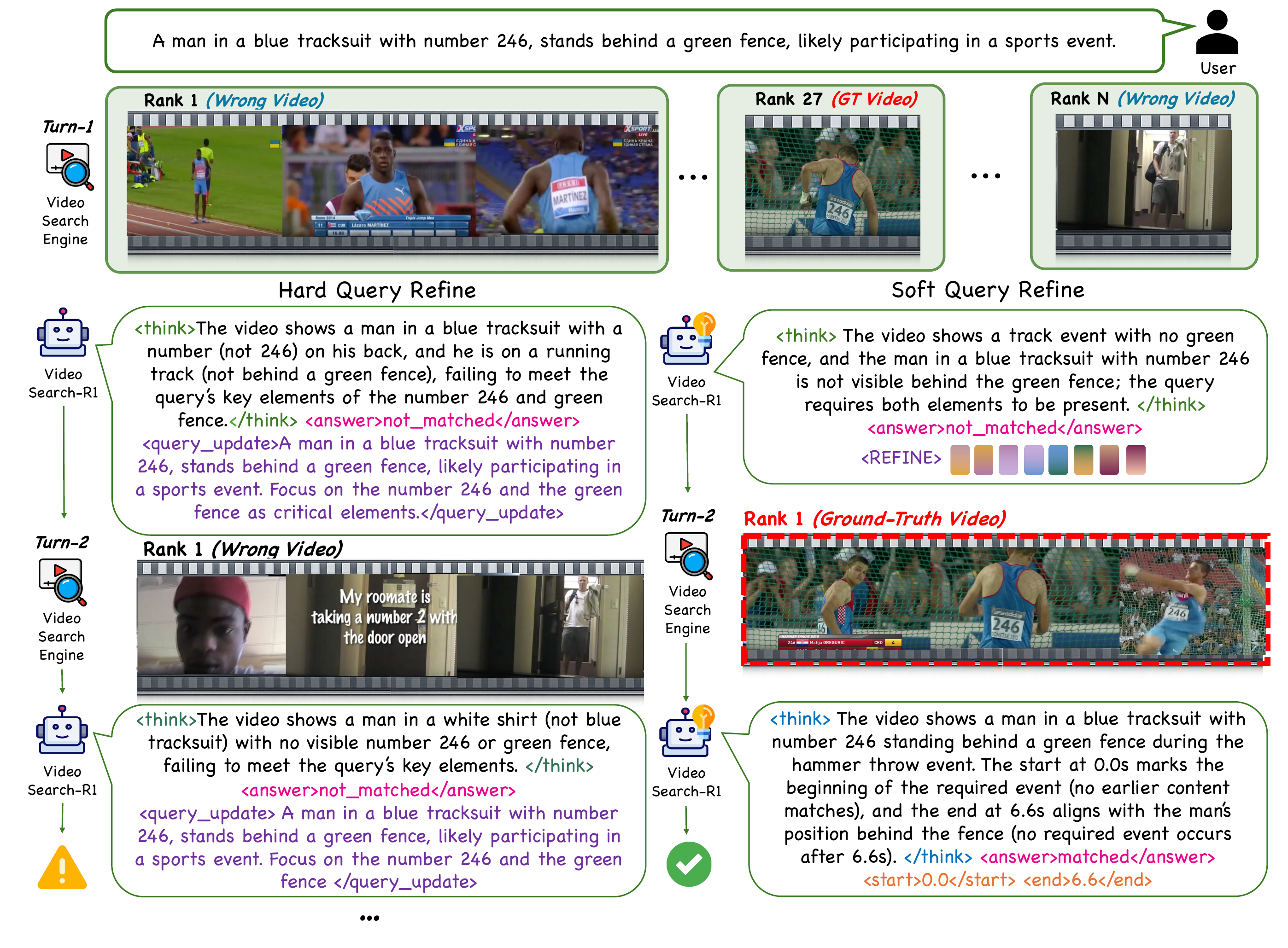}
    \caption{
        \textbf{Qualitative comparison between SQR and HQR.}
    }
    \label{fig:qual_sqr}
\end{figure}
To further analyze the necessity of continuous latent refinement, we compare SQR with hard query refinement (HQR) based on explicit text-level query refinement.
As shown in Tab.~\ref{tab:sqr}, SQR improves R@1 in VR by 7.2, compared to a 3.7 gain achieved by HQR.
This indicates that directly optimizing continuous query representations enables more fine-grained adjustments than relying on discrete query refinement.
Moreover, HQR produces substantially longer and more verbose refined queries (averaging 26.8 tokens), whereas SQR requires only eight soft query tokens to attain superior performance. 
Fig.~\ref{fig:qual_sqr} provides a qualitative example in which HQR retrieves an incorrect video, whereas SQR successfully refines the query representation and retrieves the correct video using only eight latent tokens. 
Notably, even after applying HQR, the search engine still returns the incorrect video at rank 1.
We hypothesize that the longer rewritten queries produced by HQR introduce semantic noise, which can confuse the video search engine that relies on cross-modal embedding similarity rather than strong instruction-following capabilities. 
In contrast, SQR operates directly in the continuous embedding space, enabling more precise adjustments to the query representation using only a small number of latent tokens. 
Overall, these results demonstrate that SQR refines query embeddings more efficiently and precisely than HQR, leading to improved retrieval performance while requiring significantly fewer generated tokens.

\noindent \textbf{Effect of multi-turn inference.}
Fig.~\ref{fig:turn} shows the performance trends as the number of inference turns increases.
We observe a clear improvement from the first to the second turn, after which performance saturates when $T=3$. 
This suggests that a small number of refinement turns is sufficient to effectively balance computational efficiency and retrieval accuracy.

\noindent \textbf{Qualitative results.}
Finally, we present a qualitative case study in Fig.~\ref{fig:teaser} that illustrates the multi-turn reasoning process of \textsc{VideoSearch-R1}. 
Given the textual query about `a man in a dark gray t-shirt applying lotion to a black shoe indoors', the initial retrieval at $t=1$ returns a rank-1 candidate video that shares superficial visual similarities (\eg, a man applying lotion to a shoe) but fails to capture the detailed semantics. 
During verification, the model identifies this discrepancy and generates an intermediate reasoning trace explaining the missing semantic cues (\eg, noting that the man wears a light gray t-shirt in an outdoor setting with no wooden door or framed picture), correctly predicting a \texttt{`not match'} decision followed by a \texttt{<REFINE>} token. 
Based on this mismatch, the model generates soft query tokens via SQR. 
When the search engine is re-invoked at $t=2$, the refined query representation retrieves the ground-truth video at rank-1, which was previously ranked 14. 
After confirming the match (\texttt{`match'}), the model accurately localizes the target moment (start: 0.0s, end: 9.86s) with an IoU of 0.89. 
This example demonstrates the self-correcting capability of our iterative retrieval-and-reasoning framework, which resolves earlier retrieval errors through latent-space query refinement and subsequently performs fine-grained temporal reasoning.

\section{Conclusion}
\label{sec:conclusion}

In this paper, we introduce \textsc{VideoSearch-R1}, an agentic framework that unifies inter-video retrieval and intra-video reasoning within an iterative multi-turn loop. 
The model autonomously retrieves candidate videos, verifies their semantic alignment with user intent, refines search queries, and performs reasoning grounded in the retrieved content.
We further propose Soft Query Refinement (SQR), a continuous latent-space query optimization mechanism that replaces explicit token-level rewriting. 
By avoiding verbose textual rewriting, SQR enables efficient and fine-grained query adjustments with substantially fewer generated tokens.
Trained with reinforcement learning, \textsc{VideoSearch-R1} achieves state-of-the-art performance on VCMR, demonstrating that the iterative retrieval-and-reasoning pipeline provides a robust, self-correcting foundation for complex, large-scale video understanding.
\section*{Acknowledgement}
\label{sec:Acknowledgement}

This work was supported by the InnoCORE program of the Ministry of Science and ICT (AI Meta-Scientist, N10260110, 30\%), the Electronics and Telecommunications Research Institute (ETRI) grant funded by Korean government [26ZR1200, Research on Autonomous Vision Augmentation and Extension Technologies, 30\%], and Institute of Information \& Communications Technology Planning \& Evaluation (IITP) grant funded by the Korea government (MSIT) (No. RS-2024-00443251, Accurate and Safe Multimodal, Multilingual Personalized AI Tutors, 40\%).


\bibliographystyle{splncs04}
\bibliography{main}
\clearpage
\includepdf[pages=-]{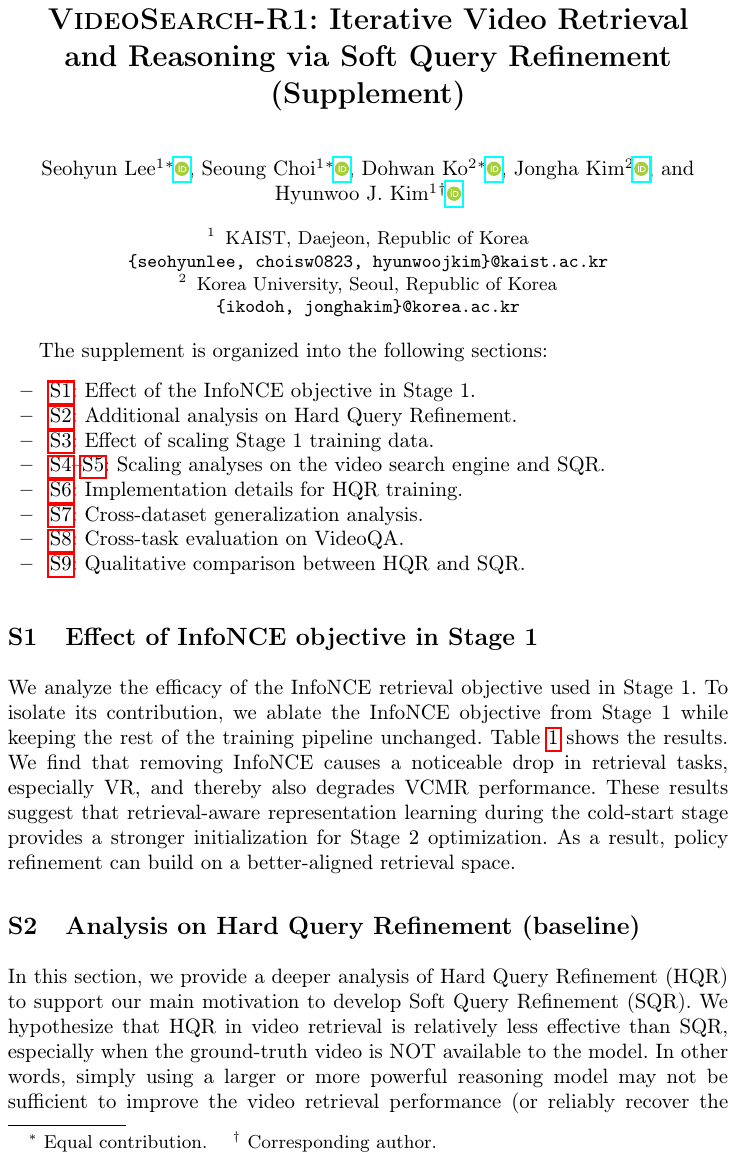}

\end{document}


\title{\texorpdfstring{
\textsc{VideoSearch-R1}: Iterative Video Retrieval \\
and Reasoning via Soft Query Refinement (Supplement)
}{
VideoSearch-R1: Iterative Video Retrieval and Reasoning via Soft Query Refinement (Supplement)
}}

\titlerunning{VideoSearch-R1 Supplement}

\author{
Seohyun Lee\inst{1}$^{*}$\orcidlink{0009-0003-4661-3486}
\and
Seoung Choi\inst{1}$^{*}$\orcidlink{0009-0006-4885-4074}
\and
Dohwan Ko\inst{2}$^{*}$\orcidlink{0000-0002-2284-1009}
\and
Jongha Kim\inst{2}\orcidlink{0009-0007-4304-941X}
\and
Hyunwoo J. Kim\inst{1}$^{\dagger}$\orcidlink{0000-0002-2181-9264}
}

\authorrunning{S. Lee et al.}

\institute{
KAIST, Daejeon, Republic of Korea\\
\email{\{seohyunlee, choisw0823, hyunwoojkim\}@kaist.ac.kr}
\and
Korea University, Seoul, Republic of Korea\\
\email{\{ikodoh, jonghakim\}@korea.ac.kr}
}

\maketitle

{\let\thefootnote\relax\footnotetext{$^{*}$ Equal contribution. \quad $^{\dagger}$ Corresponding author.}}

\setcounter{section}{0}
\renewcommand{\thesection}{S\arabic{section}}

The supplement is organized into the following sections:
\begin{itemize}
    \item ~\ref{sec:supp_analysis}: Effect of the InfoNCE objective in Stage 1.
    \item ~\ref{sec:analysis_hqr}: Additional analysis on Hard Query Refinement.
    \item ~\ref{sec:s3}: Effect of scaling Stage 1 training data.
    \item ~\ref{sec:s4}--\ref{sec:s5}: Scaling analyses on the video search engine and SQR.
    \item ~\ref{sec:s6}: Implementation details for HQR training.
    \item ~\ref{sec:supp_cross_dataset}: Cross-dataset generalization analysis.
    \item ~\ref{sec:supp_intentqa}: Cross-task evaluation on VideoQA.
    \item ~\ref{sec:s7}: Qualitative comparison between HQR and SQR.
\end{itemize}
\vspace{-0.3em}

\section{Effect of InfoNCE objective in Stage 1}
\label{sec:supp_analysis}
\begin{table*}[t]
\caption{
\textbf{Effect of removing InfoNCE loss during Stage 1.}
}
\label{tab:infonce_ablation}

\centering
\setlength{\tabcolsep}{3pt}

\begin{adjustbox}{width=\textwidth}
\begin{tabular}{l|c|ccc|c|cccc}
\toprule
\textbf{Method} & \textbf{InfoNCE} 
& \multicolumn{3}{c|}{\textbf{VCMR}}
& \textbf{VER}
& \multicolumn{4}{c}{\textbf{VR}} \\
\cmidrule(lr){3-5} \cmidrule(lr){6-6} \cmidrule(lr){7-10}

& 
& 0.3/R@1 & 0.5/R@1 & 0.7/R@1
& Acc
& R@1 & R@5 & R@10 & R@100 \\
\midrule

\textbf{VideoSearch-R1}(Stage 1)  & $\times$
& 19.6 & 17.7 & 12.4
& 66.7
& 55.6 & 78.2 & 84.9 & 96.7 \\

\rowcolor{cyan!12}
\textbf{VideoSearch-R1}(Stage 1)  & \textbf{\checkmark}
& 20.4 & 18.7 & 14.0
& 66.0
& 57.4 & 80.6 & 86.8 & 97.3 \\

\midrule


\textbf{VideoSearch-R1}(Stage 1 + Stage2)  & $\times$
& 31.2 & 28.4 & 17.6
& 74.4
& 56.3 & 78.6 & 84.9 & 96.7 \\

\rowcolor{cyan!12}
\textbf{VideoSearch-R1}(Stage 1 + Stage2)  & \textbf{\checkmark}
& \textbf{33.3} & \textbf{30.2} & \textbf{19.7}
& \textbf{74.6}
& \textbf{59.0} & \textbf{82.0} & \textbf{87.8} & \textbf{97.5} \\

\bottomrule
\end{tabular}
\end{adjustbox}
\vspace{-0.15cm}
\end{table*}
We analyze the efficacy of the InfoNCE retrieval objective used in Stage 1. 
To isolate its contribution, we ablate the InfoNCE objective from Stage 1 while keeping the rest of the training pipeline unchanged. Table~\ref{tab:infonce_ablation} shows the results. We find that removing InfoNCE causes a noticeable drop in retrieval tasks, especially VR, and thereby also degrades VCMR performance. These results suggest that retrieval-aware representation learning during the cold-start stage provides a stronger initialization for Stage 2 optimization. As a result, policy refinement can build on a better-aligned retrieval space.

\section{Analysis on Hard Query Refinement (baseline)}
\label{sec:analysis_hqr}
\begin{figure}[t]
    \centering
    \includegraphics[width=\linewidth]{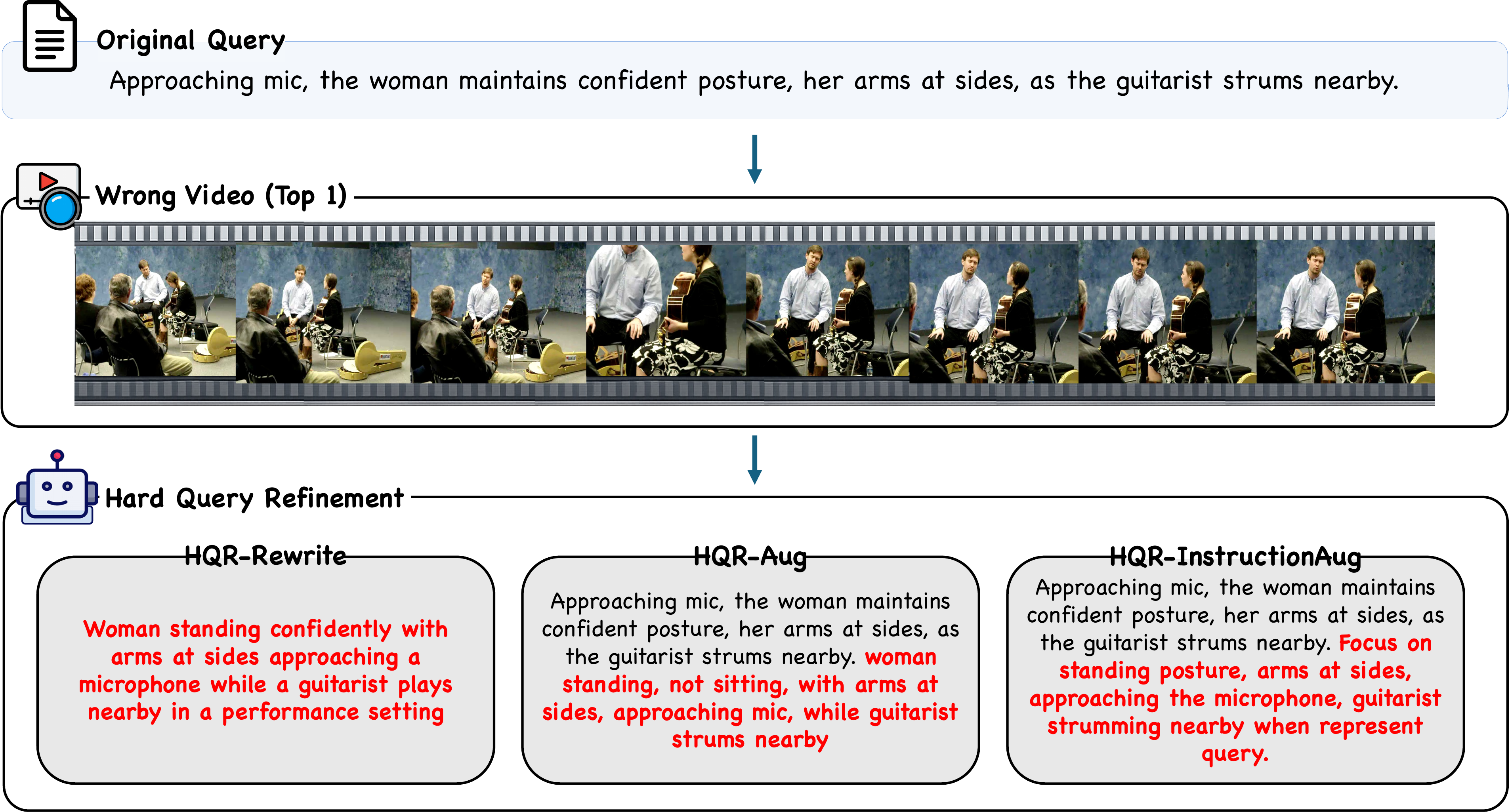}
    \caption{
        \textbf{Illustration of three HQR methods.}
    }
    \label{fig:hqr_illust}
\vspace{-0.55cm}
\end{figure}
\begin{figure}[t]
    \centering

    \begin{minipage}[t]{0.47\linewidth}
        \vspace{0pt}
        \makeatletter\def\@captype{table}\makeatother
        \centering

        \caption{
            Effect of HQR module scale in the retrieval-retry setting.
            We scale the Qwen3-VL~\cite{li2026qwen3} rewriter from 2B to 8B.
            \textit{Ours} denotes \textsc{VideoSearch-R1} with a 2B rewriter.
        }
        \label{tab:hqr_model_scale}
        \vspace{0.1cm}

        \setlength{\tabcolsep}{3.2pt}
        \renewcommand{\arraystretch}{1.05}

        \resizebox{\linewidth}{!}{%
        \begin{tabular}{@{}llrrrr@{}}
        \toprule
        Model & Size & R@1 & R@5 & R@10 & R@100 \\
        \midrule

        \multicolumn{6}{@{}l}{\textit{ActivityNet}} \\
        Qwen3-VL & 2B & 54.8 & 77.6 & 85.4 & 97.4 \\
                 & 4B & 54.9 & 76.6 & 84.3 & 97.0 \\
                 & 8B & 55.3 & 78.4 & 86.2 & 98.0 \\
        \rowcolor{oursblue}
        \textbf{Ours} & 2B & \textbf{61.1} & \textbf{81.7} & \textbf{88.5} & \textbf{98.4} \\

        \midrule
        \multicolumn{6}{@{}l}{\textit{Charades}} \\
        Qwen3-VL & 2B & 21.7 & 41.5 & 51.2 & 83.6 \\
                 & 4B & 21.8 & 39.1 & 48.2 & 81.8 \\
                 & 8B & 22.2 & 41.5 & 50.8 & 84.0 \\
        \rowcolor{oursblue}
        \textbf{Ours} & 2B & \textbf{24.6} & \textbf{42.9} & \textbf{52.2} & \textbf{84.7} \\

        \midrule
        \multicolumn{6}{@{}l}{\textit{DiDeMo}} \\
        Qwen3-VL & 2B & 55.5 & 78.8 & 85.2 & 96.6 \\
                 & 4B & 55.3 & 76.9 & 83.1 & 95.9 \\
                 & 8B & 55.6 & 79.2 & 85.6 & 97.0 \\
        \rowcolor{oursblue}
        \textbf{Ours} & 2B & \textbf{59.0} & \textbf{82.0} & \textbf{87.8} & \textbf{97.5} \\

        \bottomrule
        \end{tabular}
        }
    \end{minipage}
    \hfill
    \begin{minipage}[t]{0.51\linewidth}
        \vspace{0pt}
        \makeatletter\def\@captype{figure}\makeatother
        \centering

        \includegraphics[width=\linewidth]{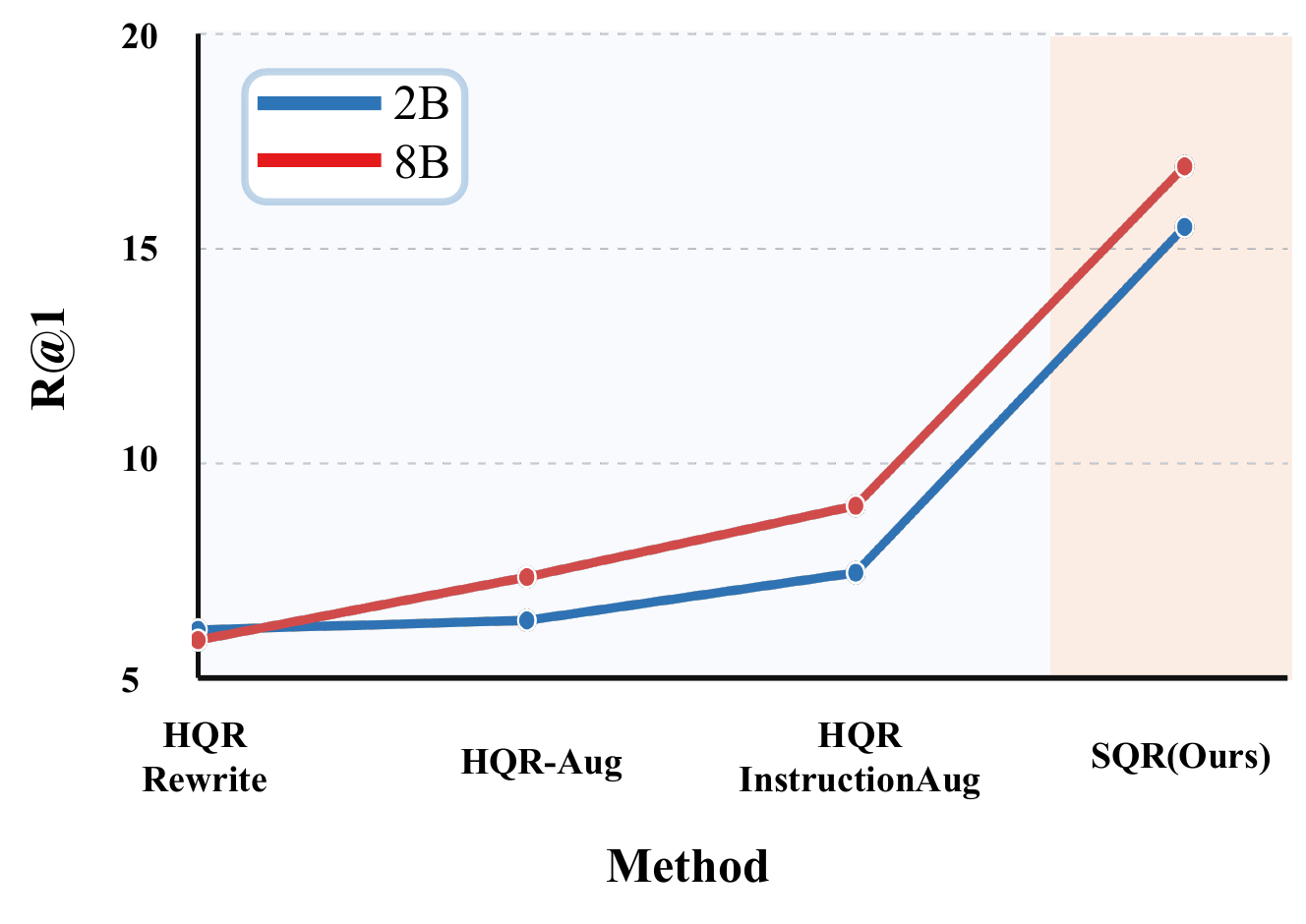}

        \caption{
            \textbf{Effect of query refinement methods on initially mismatched queries.}
            Evaluated with Qwen3-VL-Embedding-2B~\cite{li2026qwen3}
            and Qwen3-VL-Embedding-8B~\cite{li2026qwen3}.
        }
        \label{fig:HQR_METHOD}
    \end{minipage}
\end{figure}

In this section, we provide a deeper analysis of Hard Query Refinement (HQR) to support our main motivation to develop Soft Query Refinement (SQR).
We hypothesize that HQR in video retrieval is relatively less effective than SQR, especially when the ground-truth video is NOT available to the model. 
In other words, simply using a larger or more powerful reasoning model may not be sufficient to improve the video retrieval performance (or reliably recover the target video). 
To  this end, we scale the query refiner from 2B to 8B in a simple retrieval-retry setting. 
When the initial top-1 retrieval is incorrect, the model rewrites the query and re-issues the retrieval. 
As shown in Table~\ref{tab:hqr_model_scale}, retrieval performance generally improves with model size, but the gains are modest and quickly saturate. 
This result suggests that scaling reasoning capacity alone is insufficient to resolve the ambiguity of HQR.

We next examine which form of HQR is most effective for recovering the rank of the ground-truth video. To this end, we compare several HQR variants on retrieval failures, where refinement is triggered only when the initially retrieved video is incorrect.  Figure~\ref{fig:hqr_illust} illustrates the differences among the three variants in how they refine the original query. HQR-Rewrite allows the model to freely rewrite the query in order to retrieve the correct video. HQR-Aug preserves the original query and appends an additional refinement. HQR-InstructionAug explicitly distinguishes visually matched aspects from missing ones, instructing the model to suppress the former and emphasize the latter.

As shown in Fig.~\ref{fig:HQR_METHOD}, HQR-InstructionAug achieves the best performance among the HQR variants. This result suggests that effective hard query refinement is not simply a matter of rewriting the query from scratch, but of selectively redistributing semantic emphasis toward the missing aspects of the target video. In many failure cases, the retriever appears to overemphasize only part of the query, causing the ground-truth video to be ranked lower. In such cases, guiding the model to downweight the already matched aspects and upweight the missing ones is often sufficient to recover the target video. Based on this observation, we adopt HQR-InstructionAug as the default HQR design in our experiments, as it provides the strongest hard-refinement baseline among the variants we tested. Notably, even this optimized HQR setting yields only limited gains. SQR outperforms HQR not simply because the HQR baseline is less optimized, but because hard rewriting itself remains challenging when the model cannot directly access the target video. Implementation details for the HQR setup are provided in Sec.~\ref{sec:supp_implementation}.

\begin{table*}[t]
\caption{\textbf{Effect of scaling Stage 1 training data.}}

\label{tab:full_finetune}

\centering
\setlength{\tabcolsep}{3pt}

\begin{adjustbox}{width=\textwidth}
\begin{tabular}{l|ccc|c|cccc|cc}
\toprule
\textbf{Method}
& \multicolumn{3}{c|}{\textbf{VCMR}}
& \textbf{VER}
& \multicolumn{4}{c|}{\textbf{VR}}
& \textbf{\#S1 Train}
& \textbf{\#S2 Train} \\
\cmidrule(lr){2-4}
\cmidrule(lr){5-5}
\cmidrule(lr){6-9}
\cmidrule(lr){10-11}
& 0.3/R@1 & 0.5/R@1 & 0.7/R@1
& Acc
& R@1 & R@5 & R@10 & R@100
&  &  \\
\midrule

Qwen3-VL-2B\cite{bai2025qwen3} (ZS)
& 22.0 & 10.6 & 4.0
& 62.8
& 54.8 & 79.3 & 85.6 & 97.0
& -- & -- \\

\midrule

\textbf{VideoSearch-R1} (Stage 1)
& 20.4 & 18.7 & 14.0
& 66.0
& 57.4 & 80.6 & 86.8 & 97.3
& 2K & -- \\

\textbf{VideoSearch-R1} (Stage 1)
& 22.5 & 20.7 & 14.8
& 70.4
& 58.0 & 80.9 & 87.3 & \textbf{97.7}
& 22.3K & -- \\

\midrule
\rowcolor{cyan!12}
\textbf{VideoSearch-R1} (Stage 1 + Stage 2)
& \textbf{33.3} & \textbf{30.2} & \textbf{19.7}
& \textbf{74.6}
& \textbf{59.0} & \textbf{82.0} & \textbf{87.8} & 97.5
& 2K & 22.3K \\

\bottomrule
\end{tabular}
\end{adjustbox}
\end{table*}
\section{Analysis of Scaling Stage 1 Training Data}
\label{sec:s3}
We investigate whether the performance gain by Stage 2 can be obtained simply by increasing the scale of the dataset in Stage 1 by a factor of 10. 
We compare the VideoSearch-R1 pipeline in Table~\ref{tab:full_finetune} obtained by increasing the amount of Stage 1 training data. Concretely, we generate additional synthetic reasoning data with Qwen3-VL-30B~\cite{bai2025qwen3}, matching the volume of data used in Stage 2, and use them to augment Stage 1 training.
Scaling Stage 1 with additional synthetic reasoning data yields a stronger supervised baseline. However, it does not recover the gains of Stage 2. Stage 1 relies on supervised imitation and InfoNCE, requiring the model to learn retrieval, temporal grounding, and answer generation jointly. Moreover, even when the reasoning traces are generated by a much stronger model, they are not guaranteed to induce behaviors that jointly improve retrieval and temporal grounding. By contrast, Stage 2 uses reward-based optimization to let the model discover such behaviors directly from task-level signals, rather than relying solely on imitation of provided reasoning traces.

\section{Effect of Scaling the Video Search Engine}

\begin{table*}[t]
\caption{\textbf{Scaling up Video Search Engine size on DiDeMo-FIG}.}
\label{tab:searchscaleup}

\centering
\setlength{\tabcolsep}{3pt}

\begin{adjustbox}{width=\textwidth}
\begin{tabular}{c|l|ccc|c|cccc}
\toprule
\multirow{2}{*}{\makecell{\textbf{Video Search}\\\textbf{Engine Size}}}
& \multirow{2}{*}{\textbf{Method}}
& \multicolumn{3}{c|}{\textbf{VCMR}}
& \textbf{VER}
& \multicolumn{4}{c}{\textbf{VR}} \\
\cmidrule(lr){3-5} \cmidrule(lr){6-6} \cmidrule(lr){7-10}
&
& 0.3/R@1 & 0.5/R@1 & 0.7/R@1
& Acc
& R@1 & R@5 & R@10 & R@100 \\
\midrule

\multirow{2}{*}{2B}
& Qwen3-VL-2B~\cite{bai2025qwen3} (ZS) & 22.0 & 10.6 & 4.0 & 62.8 & 54.8 & 79.3 & 85.6 & 97.0 \\
& \hl{VideoSearch-R1(Stage1)} & \hl{20.4} & \hl{18.7} & \hl{14.0} & \hl{66.0} & \hl{57.4} & \hl{\textbf{80.6}} & \hl{86.8} & \hl{97.3} \\
\midrule

\multirow{2}{*}{8B}
& Qwen3-VL-2B~\cite{bai2025qwen3} (ZS) & 23.6 & 11.6 & 4.5 & 65.9 & 61.3 & \textbf{84.5} & \textbf{90.0} & 98.0 \\
& \hl{VideoSearch-R1(Stage1)}  & \hl{\textbf{26.0}} & \hl{\textbf{24.0}} & \hl{\textbf{17.4}} & \hl{\textbf{70.0}} & \hl{\textbf{63.5}} & \hl{84.3} & \hl{\textbf{90.0}} & \hl{\textbf{98.3}} \\

\bottomrule
\end{tabular}
\end{adjustbox}

\end{table*}
\label{sec:s4}
We investigate whether the limited effectiveness of HQR is due to the video search engine not adequately responding to textual refinement instructions. For this analysis, we scale the search engine from Qwen3-VL-Embedding-2B~\cite{li2026qwen3} to Qwen3-VL-Embedding-8B~\cite{li2026qwen3} and evaluate both zero-shot retrieval and Stage 1 training.
Table~\ref{tab:searchscaleup} shows that the 8B search engine is substantially stronger than the 2B model across zero-shot retrieval and Stage 1 evaluation. While query refinement remains beneficial even with the stronger 8B search engine, Fig.~\ref{fig:HQR_METHOD} shows that scaling the search engine does not qualitatively change the behavior of HQR, and the gains from hard query refinement remain limited.
These observations suggest that improving the video search engine alone is insufficient to fully address the limitation of HQR, which further motivates our use of soft query refinement.

\section{Effect of Scaling the SQR Module}
\begin{table*}[t]
\caption{\textbf{Effect of scaling the VideoSearch-R1 on DiDeMo-FIG.}}
\label{tab:sqrscaleup}

\centering
\setlength{\tabcolsep}{4pt}

\begin{adjustbox}{width=\textwidth}
\begin{tabular}{l c | ccc | c | cccc}
\toprule
\multirow{2}{*}{\textbf{Method}} 
& \multirow{2}{*}{\textbf{Size}}
& \multicolumn{3}{c|}{\textbf{VCMR}}
& \textbf{VER}
& \multicolumn{4}{c}{\textbf{VR}} \\
\cmidrule(lr){3-5} \cmidrule(lr){6-6} \cmidrule(lr){7-10}
& 
& 0.3/R@1 & 0.5/R@1 & 0.7/R@1
& Acc
& R@1 & R@5 & R@10 & R@100 \\
\midrule

\multirow{2}{*}{Qwen3-VL~\cite{bai2025qwen3}(ZS)}
& 2B & 22.0 & 10.6 & 4.0 & 62.8 
& \multirow{2}{*}{54.8} & \multirow{2}{*}{79.3} & \multirow{2}{*}{85.6} & \multirow{2}{*}{97.0} \\
& 4B & 16.6 & 9.5 & 4.0 & 69.2 
&  &  &  &  \\
\midrule

\multirow{2}{*}{\textbf{VideoSearch-R1 (Stage1)}}
& 2B & 20.4 & 18.7 & 14.0 & 66.0
& 57.4 & \textbf{80.6} & 86.8 & 97.3 \\
& 4B & \textbf{27.2} & \textbf{25.3} & \textbf{19.0} & \textbf{73.7}
& \textbf{58.6} & \textbf{80.6} & \textbf{87.2} & \textbf{97.9} \\

\bottomrule
\end{tabular}
\end{adjustbox}
\end{table*}
\label{sec:s5}
We next examine the effect of scaling the SQR module. As shown in Table~\ref{tab:sqrscaleup}, scaling the SQR module from 2B to 4B consistently improves performance across VCMR, VER, and VR. These results indicate that increasing the capacity of the soft refinement module improves the overall retrieval pipeline.
While the advantage of the 4B model is less evident in the zero-shot setting, likely because it behaves more conservatively, the improvement becomes much clearer after task-specific training. This suggests that the larger SQR module learns a more effective refinement policy, leading to stronger overall performance.

\section{Implementation Details for HQR Training}
\label{sec:supp_implementation}
\label{sec:s6}
As discussed in Sec.~\ref{sec:analysis_hqr}, HQR lacks a unique ground-truth rewritten query for supervision. Instead, we rely on the hard-query rewriting scheme that achieved the strongest retrieval improvement in our analysis. Using Qwen3-VL-30B~\cite{bai2025qwen3}, we generate refined queries with reasoning path and use them as cold-start targets for HQR training.
For a fair comparison, we keep the overall training setup identical to SQR, except that HQR uses explicit textual query rewrites instead of soft query tokens. In Stage 1, the InfoNCE objective used in SQR is not available for HQR, since the refinement is produced in discrete text space. We therefore use supervised fine-tuning with next-token prediction on the constructed rewrite targets. We use the same reasoning, verification, and grounding supervision as in SQR. In Stage 2, we follow the same GRPO reward design and hyperparameter setting as in SQR. Retrieval rewards are computed in the same way, except that retrieval is performed with the rewritten text query in HQR rather than with soft query tokens as in SQR.
\section{Cross-dataset Generalization}
\label{sec:supp_cross_dataset}
\begin{table}[t]
\centering
\caption{Cross-dataset generalization performance. VideoSearch-R1 is trained on one dataset and evaluated on another without target-dataset fine-tuning.}
\label{tab:cross_dataset_generalization}
\resizebox{\linewidth}{!}{
\begin{tabular}{lccccc c cccc}
\toprule
\multirow{2}{*}{Method} & \multirow{2}{*}{Train} & \multirow{2}{*}{Test}
& \multicolumn{3}{c}{VCMR} & \multicolumn{1}{c}{VER} & \multicolumn{4}{c}{VR} \\
\cmidrule(lr){4-6} \cmidrule(lr){7-7} \cmidrule(lr){8-11}
& & & 0.3/R@1 & 0.5/R@1 & 0.7/R@1 & Acc & R@1 & R@5 & R@10 & R@100 \\
\midrule
Qwen3-VL-2B (ZS) & -- & DiDeMo-FIG
& 22.0 & 10.6 & 4.0 & 62.8 & 54.8 & 79.3 & 85.6 & \textbf{97.0} \\
\rowcolor{LightCyan}
\textbf{VideoSearch-R1} & Charades-FIG & DiDeMo-FIG
& \textbf{26.1} & \textbf{17.3} & \textbf{7.1} & \textbf{65.8} & \textbf{57.2} & \textbf{80.4} & \textbf{86.6} & \textbf{97.0} \\
\midrule
Qwen3-VL-2B (ZS) & -- & Charades-FIG
& 12.2 & 7.2 & 2.9 & 30.0 & 21.6 & 41.8 & 51.5 & 84.2 \\
\rowcolor{LightCyan}
\textbf{VideoSearch-R1} & DiDeMo-FIG & Charades-FIG
& \textbf{15.1} & \textbf{9.3} & \textbf{4.0} & \textbf{51.3} & \textbf{23.4} & \textbf{43.6} & \textbf{53.4} & \textbf{84.9} \\
\bottomrule
\end{tabular}
}
\end{table}
\noindent{VideoSearch-R1 learns transferable query refinement rather than relying only on dataset-specific retrieval patterns.}
To evaluate cross-dataset transfer, we train VideoSearch-R1 on one video corpus and evaluate it on another corpus without target-dataset fine-tuning.
As shown in Table~\ref{tab:cross_dataset_generalization}, VideoSearch-R1 consistently improves over the zero-shot Qwen3-VL-2B baseline in both transfer directions.
Notably, it improves VCMR 0.5/R@1 from 10.6 to 17.3 when transferring from Charades-FIG to DiDeMo-FIG, and from 7.2 to 9.3 when transferring from DiDeMo-FIG to Charades-FIG.
The corresponding gains in VR further indicate that the refinement policy remains effective across different video corpora.
These results suggest that VideoSearch-R1 does not merely memorize corpus-specific biases, but transfers its iterative verification and refinement behavior across datasets.
Instead, VideoSearch-R1 can transfer its iterative verification and refinement behavior across different video corpora.
\section{Cross-task Evaluation: VideoQA}
\label{sec:supp_intentqa}

\providecommand{\gain}[1]{\hspace{2pt}{\scriptsize\textcolor{black!55}{$\uparrow$#1}}}

\begin{table}[t]
\centering
\caption{Cross-task evaluation on IntentQA. We report VideoQA accuracy and video retrieval performance.}
\label{tab:intentqa}
\resizebox{0.82\linewidth}{!}{
\begin{tabular}{lcc cccc}
\toprule
\multirow{2}{*}{Method} & \multirow{2}{*}{Size} & \multirow{2}{*}{VQA Acc.} & \multicolumn{4}{c}{VR} \\
\cmidrule(lr){4-7}
& & & R@1 & R@5 & R@10 & R@100 \\
\midrule
Qwen3-VL~\cite{bai2025qwen3}
& 2B & 39.4 & 29.5 & 49.7 & 59.9 & 88.0 \\
\rowcolor{LightCyan}
\textbf{VideoSearch-R1}
& 2B
& \textbf{57.3}\gain{17.9}
& \textbf{34.2}\gain{4.7}
& \textbf{55.4}\gain{5.7}
& \textbf{65.0}\gain{5.1}
& \textbf{92.1}\gain{4.1} \\
\bottomrule
\end{tabular}
}
\end{table}

\noindent{VideoSearch-R1 extends beyond temporal grounding to retrieval-conditioned VideoQA.}
To evaluate whether the proposed iterative retrieval-and-reasoning framework can transfer to a different intra-video reasoning task, we conduct an additional evaluation on IntentQA\cite{li2023intentqa} .
In this setting, the model first retrieves a relevant video from the corpus and then answers the given question based on the retrieved content. As shown in Table~\ref{tab:intentqa}, VideoSearch-R1 improves VideoQA accuracy from 39.4 to 57.3 over the Qwen3-VL-2B baseline.
It also improves retrieval performance, increasing VR R@1 from 29.5 to 34.2 and R@100 from 88.0 to 92.1.
These results suggest that the learned verification and query refinement policy can benefit retrieval-conditioned video reasoning beyond temporal grounding.


\clearpage
\section{Comparison between HQR and SQR: A Case Study}
\label{sec:s7}
\begin{figure}[!ht]
\vspace{-0.6cm}
    \centering
    \includegraphics[width=\linewidth]{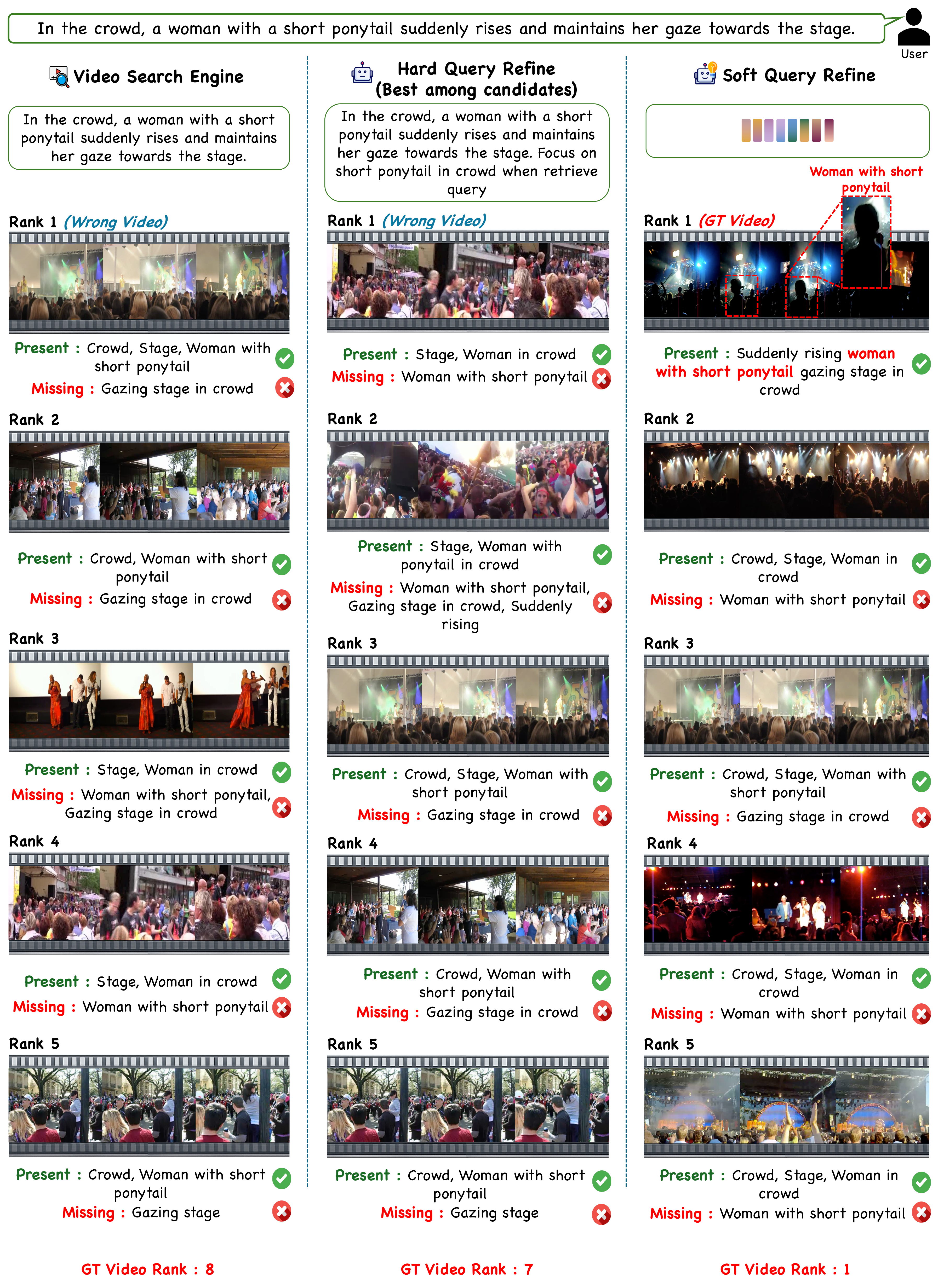}
    \caption{
        \textbf{Qualitative comparison between SQR and HQR.}
    }
    \label{fig:hqr_qual}
\end{figure}
\begin{figure}[t]
    \centering
    \includegraphics[width=\linewidth]{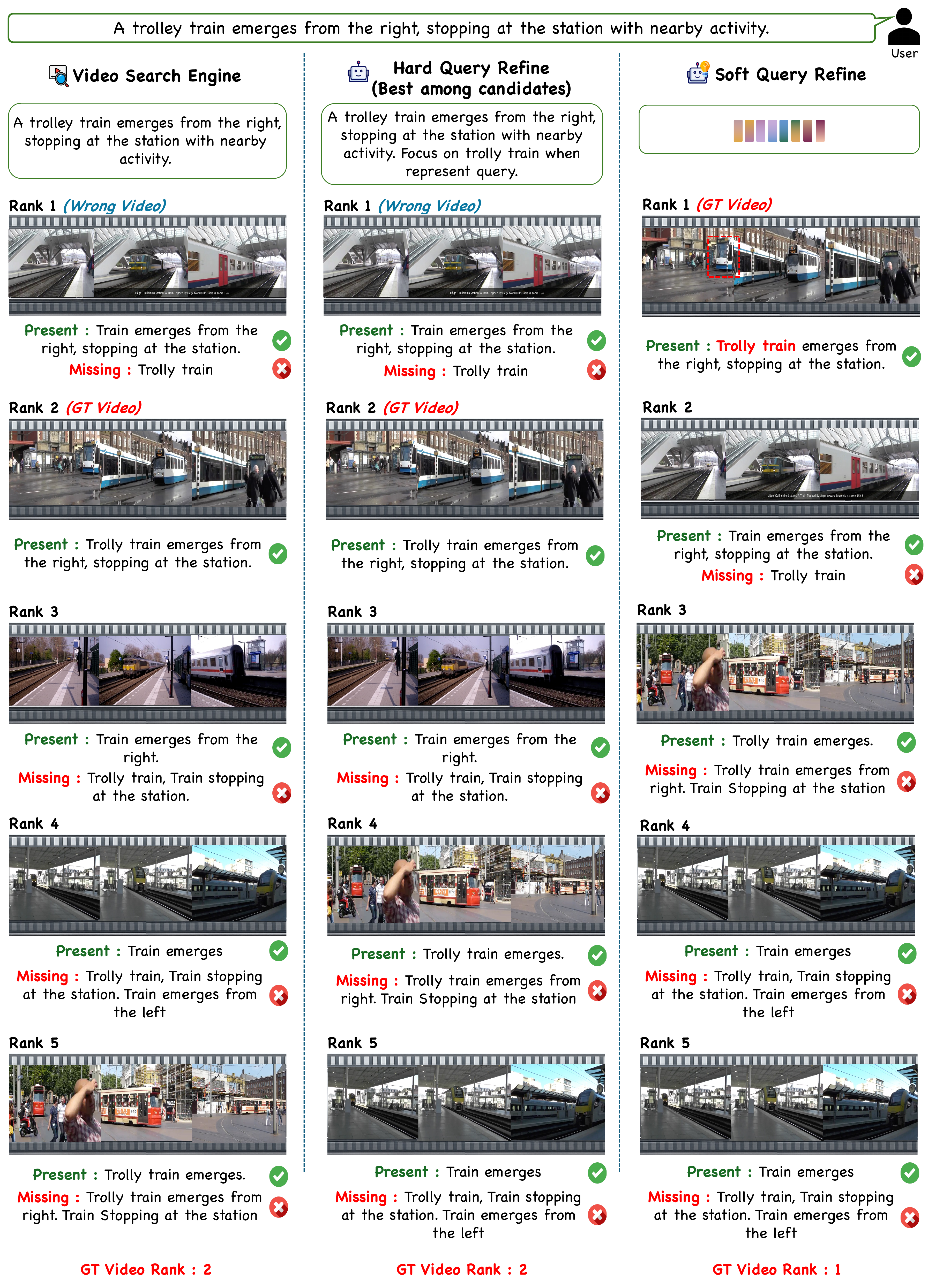}
    \caption{
        \textbf{Qualitative comparison between SQR and HQR.}
    }
    \label{fig:hqr_qual2}
\end{figure}
\begin{figure}[t]
    \centering
    \includegraphics[width=\linewidth]{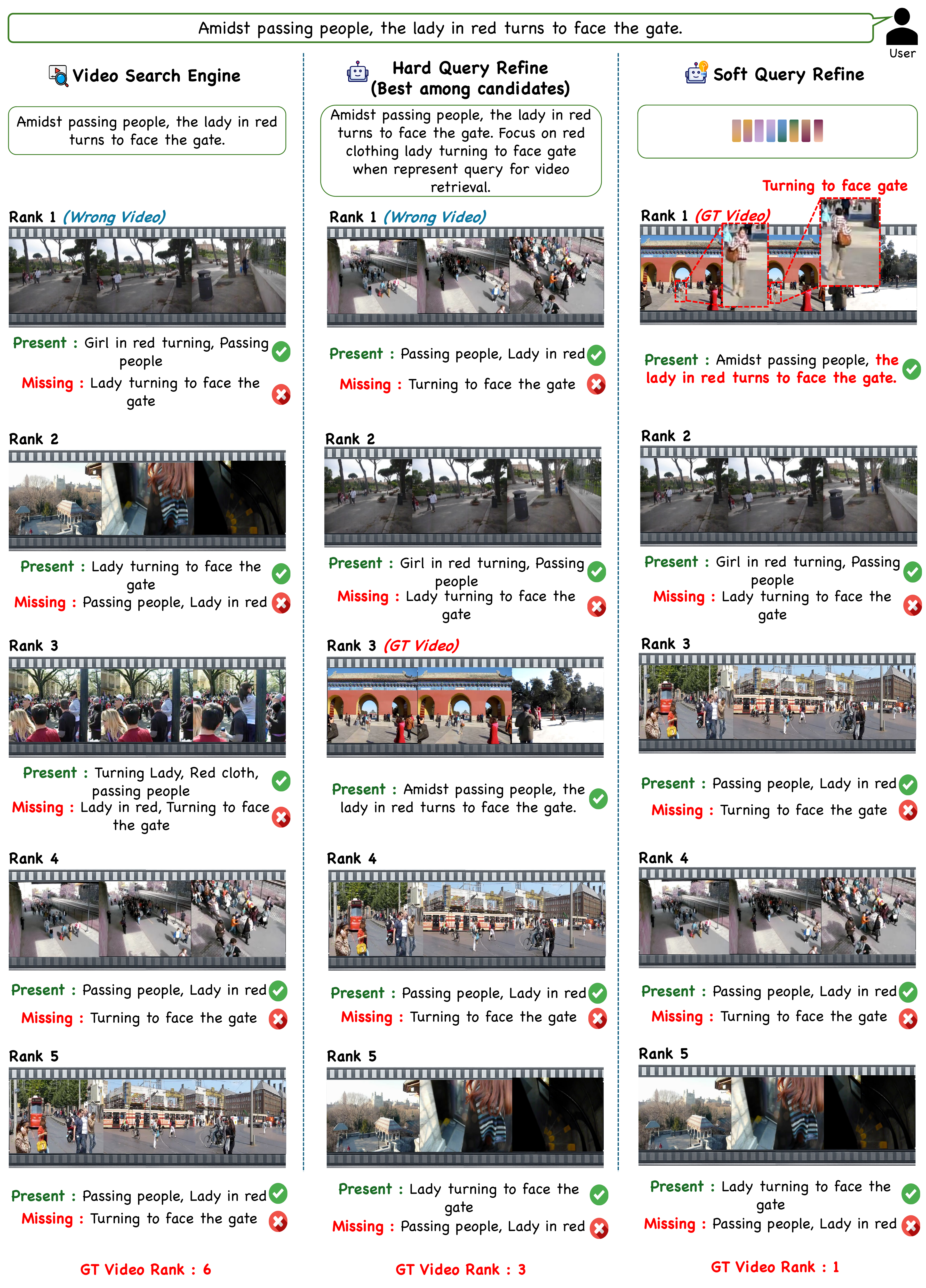}
    \caption{
        \textbf{Qualitative comparison between SQR and HQR.}
    }
    \label{fig:hqr_qual3}
\end{figure}
\begin{figure}[t]
    \centering
    \includegraphics[width=\linewidth]{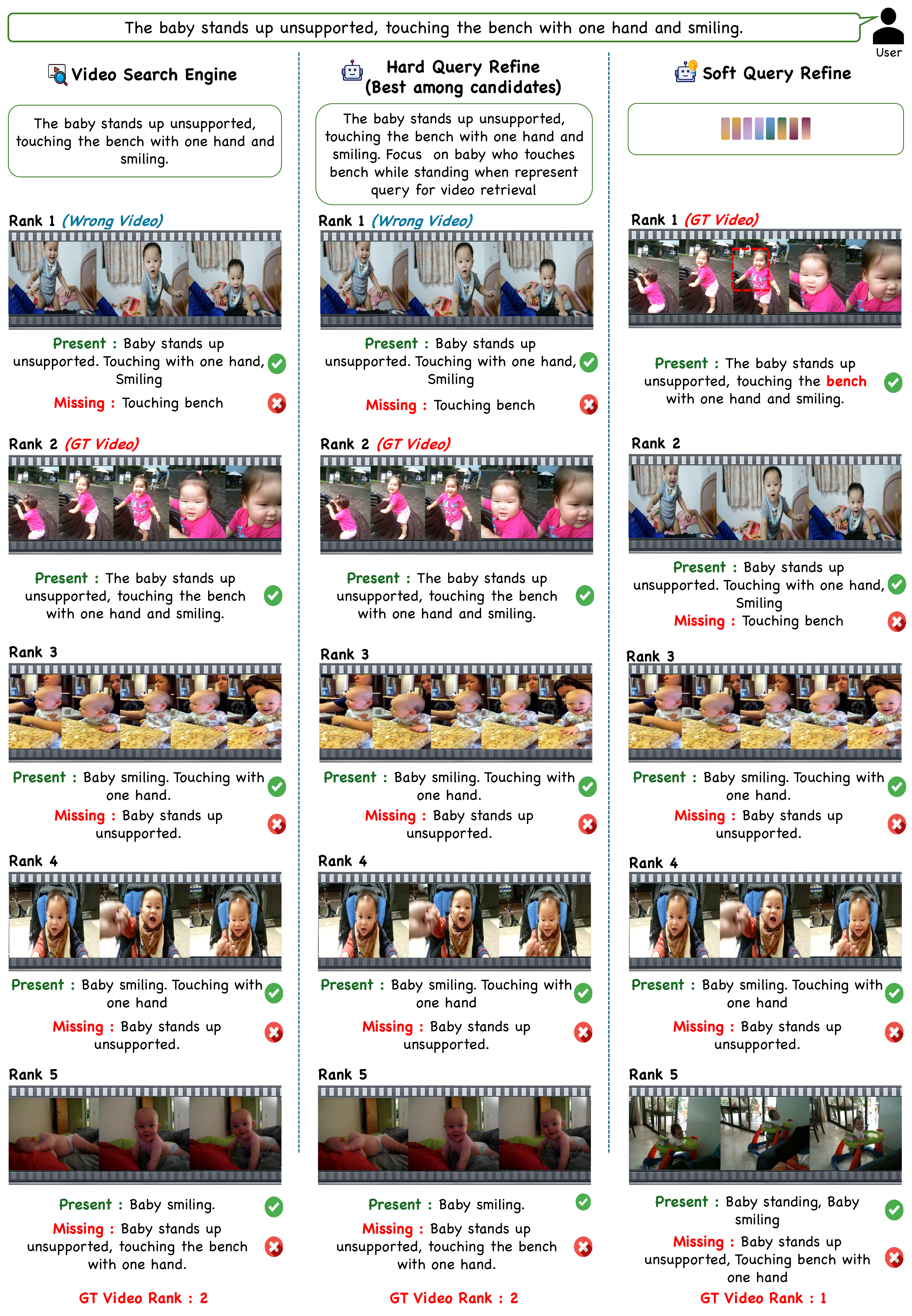}
    \caption{
        \textbf{Qualitative comparison between SQR and HQR.}
    }
    \label{fig:hqr_qual4}
\end{figure}
\begin{figure}[t]
    \centering
    \includegraphics[width=\linewidth]{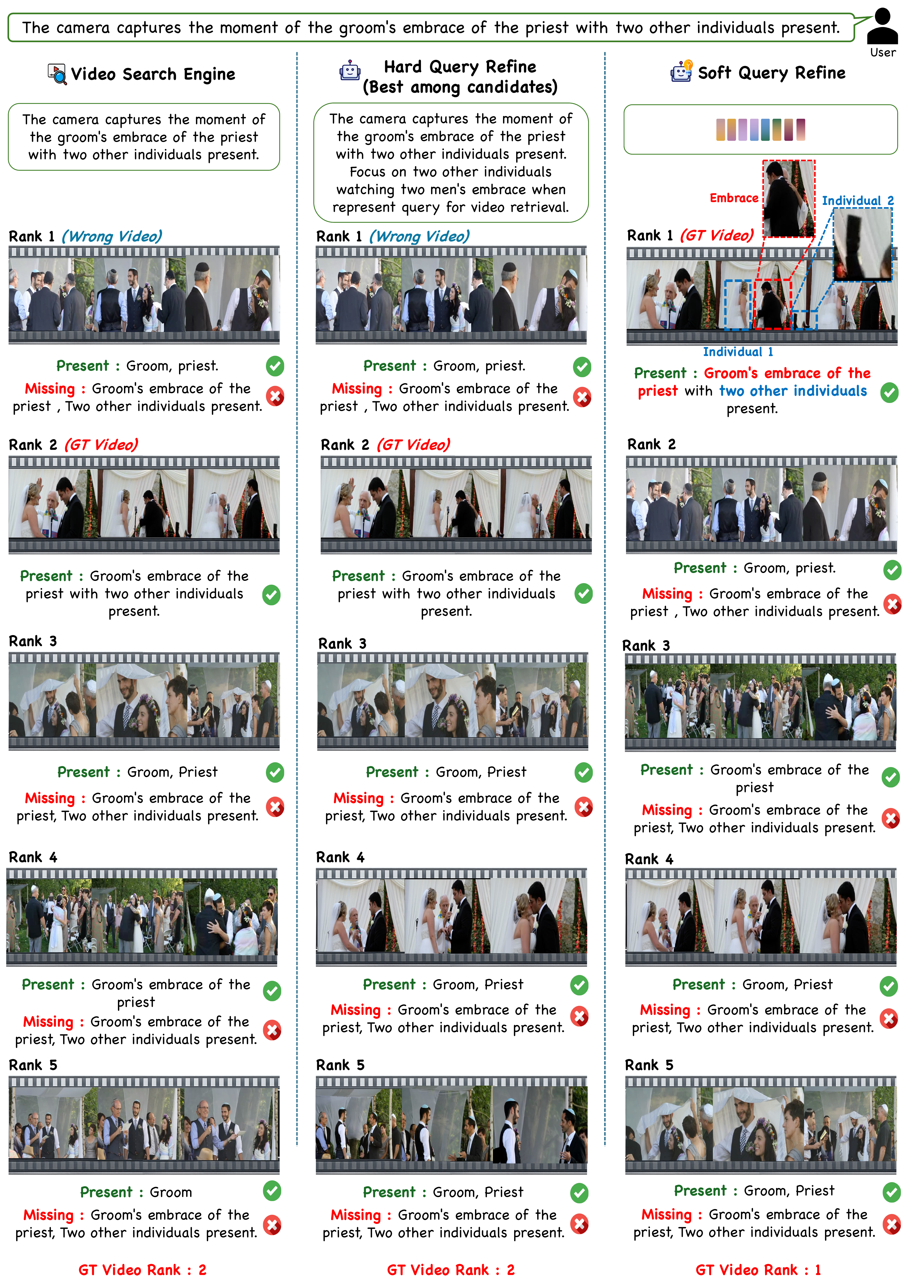}
    \caption{
        \textbf{Qualitative comparison between SQR and HQR.}
    }
    \label{fig:hqr_qual5}
\end{figure}

\clearpage
\bibliographystyle{splncs04}
\bibliography{supple}